\documentclass[lettersize,journal]{IEEEtran}
\usepackage{amsmath,amsfonts}
\usepackage[linesnumbered,ruled,vlined]{algorithm2e}
\usepackage{array}
\usepackage[caption=false,font=normalsize,labelfont=sf,textfont=sf]{subfig}
\usepackage{textcomp}
\usepackage{stfloats}
\usepackage[pagebackref,breaklinks,colorlinks]{hyperref}
\usepackage{verbatim}
\usepackage[pdftex]{graphicx}
\usepackage[caption=false,font=normalsize,labelfont=sf,textfont=sf]{subfig}
\usepackage{cite}
\usepackage{arevmath}     
\usepackage[noend]{algpseudocode}
\usepackage{amssymb}
\usepackage{booktabs}
\usepackage{multirow}
\usepackage{colortbl}
\usepackage{hhline}
\newcommand{\sign}[1]{\mathrm{sign}(#1)}
\newcommand{\dis}[2]{\mathrm{dis}(#1,#2)}

\usepackage[capitalize]{cleveref}
\crefname{section}{Sec.}{Secs.}
\Crefname{section}{Section}{Sections}
\Crefname{table}{Table}{Tables}
\crefname{table}{Tab.}{Tabs.}
\crefname{equation}{Eq.}{Eqs.}

\begin{document}
\title{Vision Transformer Based Video Hashing Retrieval \\for Tracing the Source of Fake Videos}

\author{Pengfei Pei, Xianfeng Zhao, Yun Cao, Jinchuan Li, and Xuyuan Lai
\thanks{Pengfei Pei, Xianfeng Zhao (corresponding author), Yun Cao, Jinchuan Li, and Xuyuan Lai are with State Key Laboratory of Information Security, Institute of Information Engineering, Chinese Academy of Sciences, Beijing, China. (e-mail: \{peipengfei,zhaoxianfeng,caoyun,lijinchuan,laixuyuan\}@iie.ac.cn).}
\thanks{Pengfei Pei, Xianfeng Zhao, Yun Cao, Jinchuan Li, and Xuyuan Lai are also with School of Cyber Security, University of Chinese Academy of Sciences, Beijing, China}
}

\markboth{Journal of \LaTeX\ Class Files,~Vol.~14, No.~8, August~2021}
{Shell \MakeLowercase{\textit{et al.}}: A Sample Article Using IEEEtran.cls for IEEE Journals}

\maketitle

\begin{abstract}
In recent years, the spread of fake videos has brought great influence on individuals and even countries. It is important to provide robust and reliable results for fake videos. The results of conventional detection methods are not reliable and not robust for unseen videos. Another alternative and more effective way is to find the original video of the fake video. For example, fake videos from the Russia-Ukraine war and the Hong Kong law revision storm are refuted by finding the original video. We use an improved retrieval method to find the original video, named ViTHash. Specifically, tracing the source of fake videos requires finding the unique one, which is difficult when there are only small differences in the original videos. To solve the above problems, we designed a novel loss Hash Triplet Loss. In addition, we designed a tool called Localizator to compare the difference between the original traced video and the fake video.  We have done extensive experiments on FaceForensics++, Celeb-DF and DeepFakeDetection, and we also have done additional experiments on our built three datasets: DAVIS2016-TL (video inpainting), VSTL (video splicing) and DFTL (similar videos). Experiments have shown that our performance is better than state-of-the-art methods, especially in cross-dataset mode. Experiments also demonstrated that ViTHash is effective in various forgery detection: video inpainting, video splicing and deepfakes. Our code and datasets have been released on GitHub: \url{https://github.com/lajlksdf/vtl}.
\end{abstract}

\begin{IEEEkeywords}
Video forensics, tracing the source, video hashing retrieval, vision transformer.
\end{IEEEkeywords}
\section{Introduction}
\label{sec:intro}
Since DeepFakes \cite{DeepfakesMTD,dp1,dp2} and object-based \cite{ImageTampering,ob1,ob2} tampering have become a global phenomenon, there has been an increasing interest in forgery detection.
When created with malicious intent, they have the potential to intimidate individuals and organizations.
These fake videos can lead to serious political, social, financial or legal consequences.
When we need some material as legal evidence, the results of traditional detection as a probability value is not reliable.
The hot events such as the Russian-Ukraine war also turns out that finding the original video is the most effective way to rumor.

\begin{figure}[t]
    \centering
    \includegraphics[width=1\linewidth]{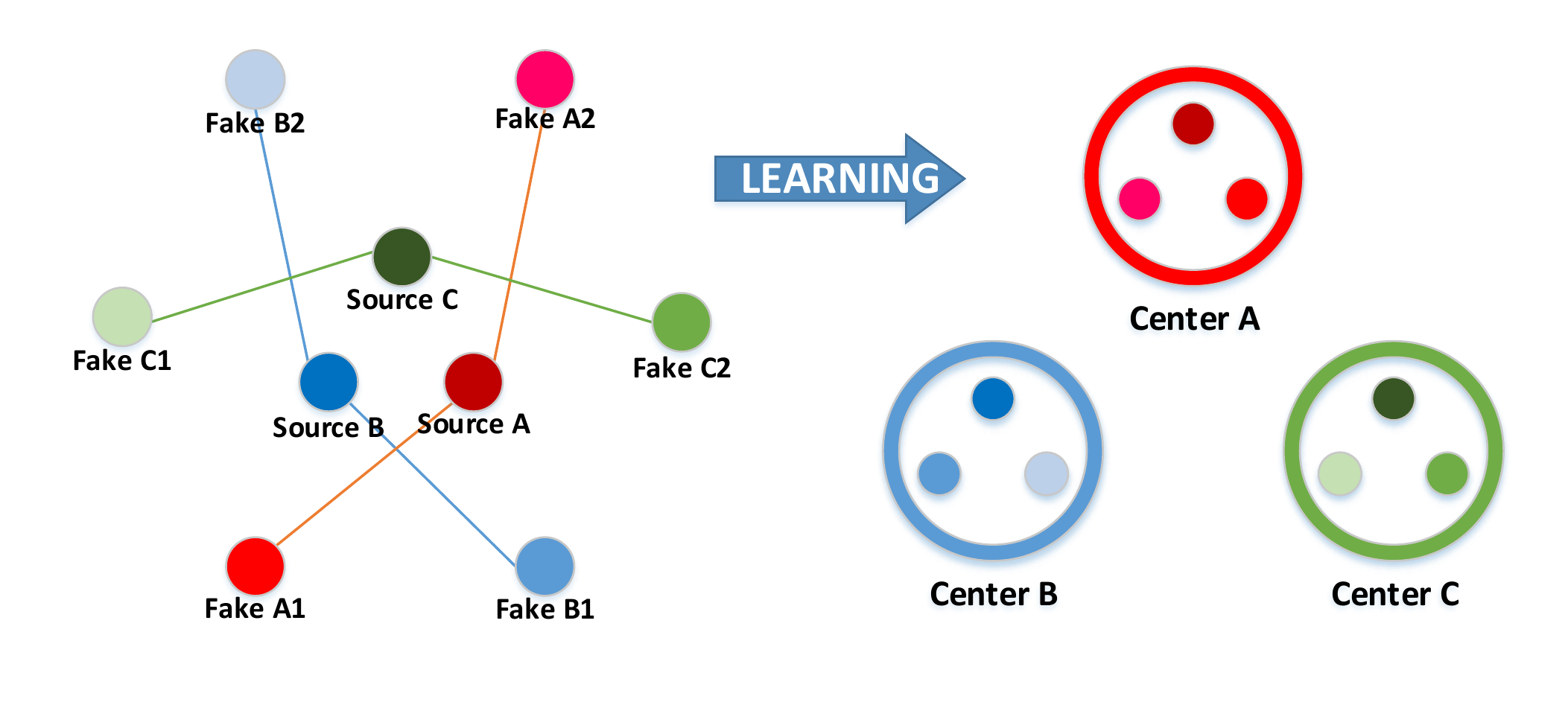}
    \caption{
        Illustration of the Hash Triplet Loss learning binary hash centers. The all samples of the set $\mathbb{X} \in \left \{ SourceX, FakeX_{1}, FakeX_{2}, \dots,  FakeX_{n}\right \}$ is a group.
        The triplet samples $(SourceX, FakeX_{i}, FakeX_{j})$ as a unit to learn $CenterX$, and the unit always contains $SourceX$.
        The \textit{Hash Triplet Loss} aims to widen the Hamming distance between each group, and lessen the Hamming distance between the original video and related fake videos.
    }
    \label{fig:htl_learning}
\end{figure}

In this paper, we proposed a vision transformer (ViT) based video hashing retrieval method to trace the source video of fake videos named ViTHash. Hashing retrieval has been widely used in search technology. One method is to generate hash centers after training. The method \cite{CSQ} trains a hash center that is closest to all samples, which is similar to clustering methods. Another method is to generate Near-Optimal hash centers before training, and the Hamming distance between these hash centers is close to or equal to half of the hash bits\cite{TBH}. When training, make the sample close to the hash center as much as possible. However, the pre-trained hash centers have the potential to destroy the internal similarity of the original data. 

\begin{figure*}[ht]
    \centering
    \includegraphics[width=.9\textwidth]{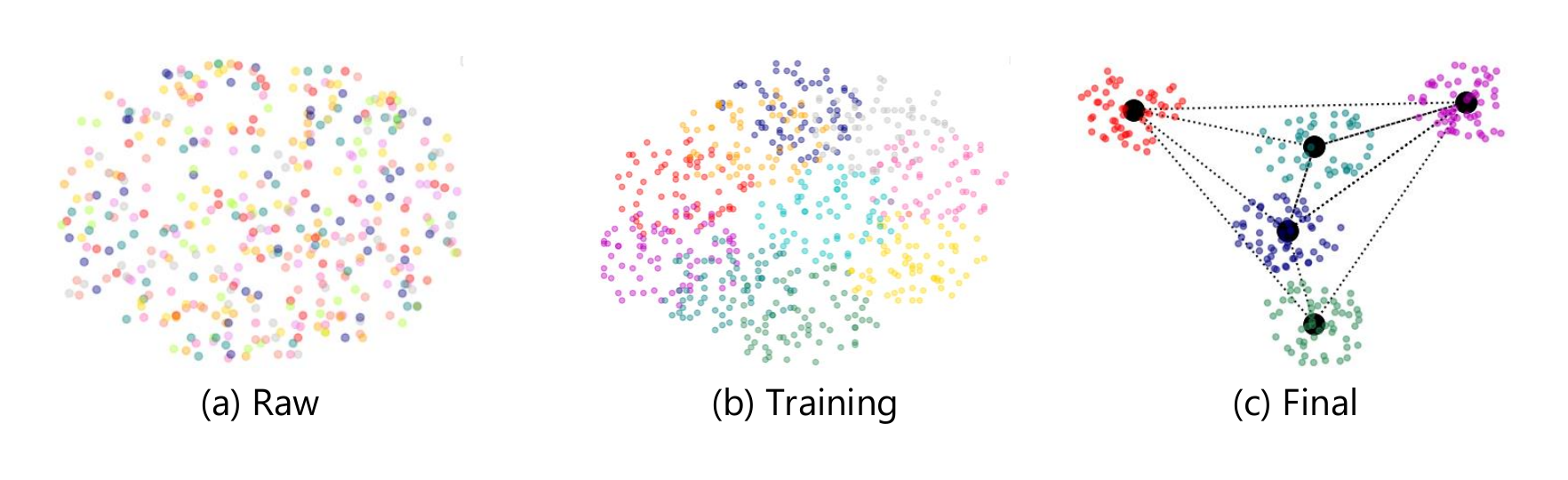}
    \caption{
        The hash distribution of learning Hash Centers. \textbf{(a)} Before training, the hash distribution of dataset is scattered. \textbf{(b)} In training, the hash distribution of each group in the dataset is gradually clustered, and the hash centers of dataset are changing. \textbf{(c)} Finally, the hash distribution of the dataset is sparse, the Hamming distance between the hash distributions of each group of data is very small, and the average Hamming distance between groups is close to half of the hash bits. The hash centers (black points) are far away from each other, and each group of data around a hash center.
    }
    \label{fig:htl_data_dist}
\end{figure*}

As shown in Fig.~\ref{fig:htl_learning} and Fig.~\ref{fig:htl_data_dist}, the challenge of using the video hashing method to trace the source is that the hash binary codes of videos are intertwined and difficult to discrimination. The traditional hashing retrieval methods are based on classification \cite{TBH,CSQ}, which are usually used to find similar images or videos, such as cars and cats, and to pursue the accuracy of Top-N. The sources tracing needs to find the original video of the fake video and pursue the accuracy of Top-1. To solve the above problems, we designed the Hash Triplet Loss. Hash Triplet Loss takes the original video as a center, which makes all samples close to the original video, and makes the Hamming distance of different hash centers as far as possible. Hash Triplet Loss uses three videos as a unit to dynamically train hash centers. Each unit of videos contains the original video and two random fake videos. As a result, the trained hash center of all fake videos in one group will be around the original video. Our main contributions are as follows:
\begin{itemize}
    \item \textbf{Novel Architecture Design}. We designed a novel architecture to detect fake videos by tracing the source of the fake video, which can provide an irrefutable proof rather than output a possible value.
    \item \textbf{Hash Triplet Loss}. We designed a new loss for our proposed method, which helps us better distinguish between the original video and similar videos with only minor differences.
    \item \textbf{New Datasets}. Due to the lack of relevant object forgery video datasets, we built three datasets to verify the performance of our method in different forgery scenes.
\end{itemize}

\begin{figure*}[ht]
    \centering
    \includegraphics[width=1\textwidth]{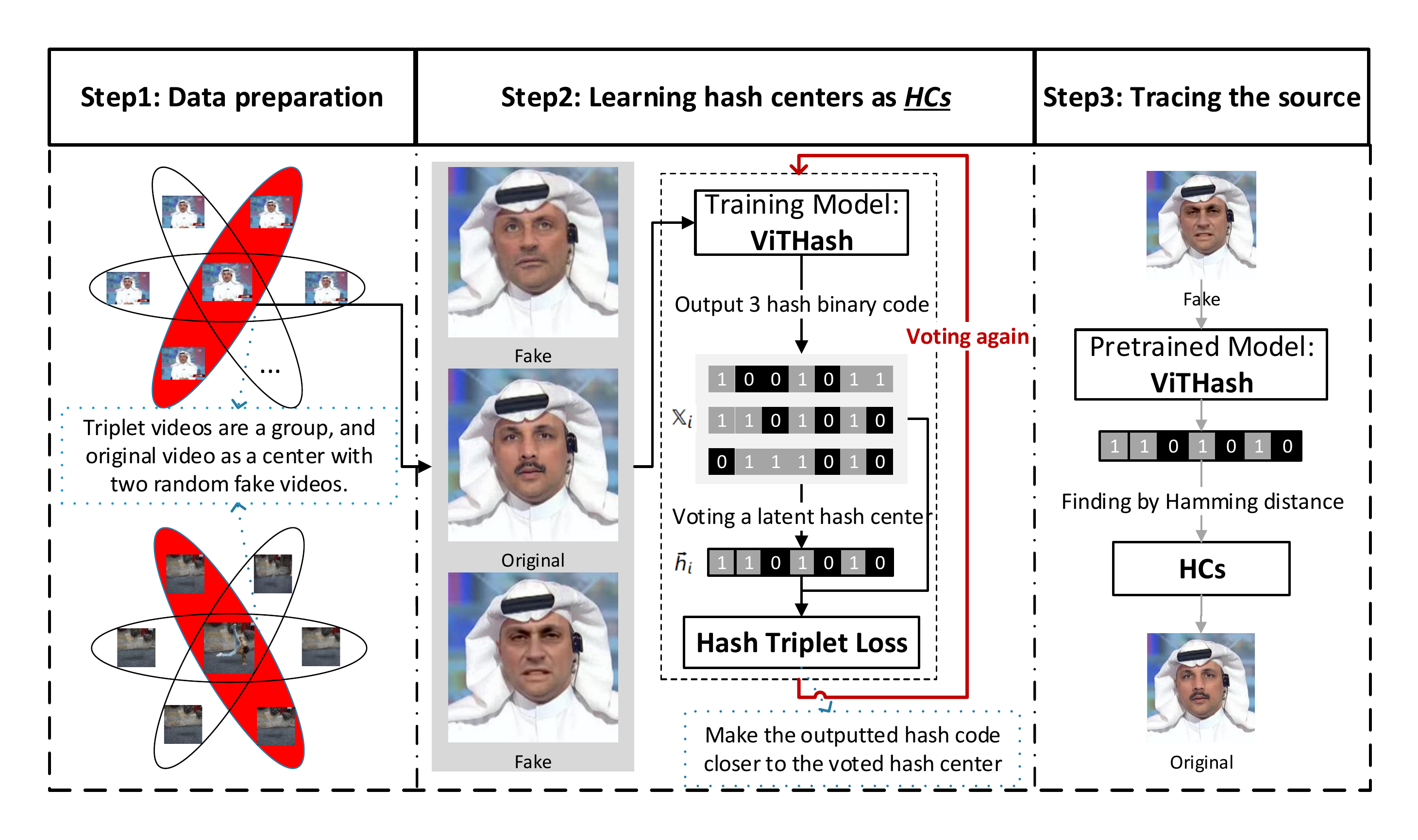}
    \caption{
        The process of training hash centers and tracing the source. \textbf{(Step1)} Prepare the training data: The original video and the related fake videos as a group, and each group trains a hash center. \textbf{(Step2)} Training hash centers: Dynamically training hash centers, the Hamming distance between hash centers is increasing with training iterations. The hash center will be re-voted after each training iteration. The Hamming distance between the re-voted hash center and the hash binary codes of the group's data is smaller. \textbf{(Step3)} Tracing the source of fake videos: The fake video is inputted into ViTHash and output a hash binary code. The nearest original video will be found by the hash binary code of the fake video.
    }
    \label{fig:flowchart}
\end{figure*}

\section{Related Works}
\label{sec:related}

\subsection{DeepFake}
In recent years, many DeepFakes related detection methods have been proposed. The research on DeepFakes mainly focuses on using binary classification method to identify the real/fake of videos. DeepFake detection methods have two main categories. The methods in the first category are based on inconsistent artifacts~\cite{DeepfakeLip,DeepfakeInConsistency}.
In the early, the DeepFakes with such low visual qualities can be clearly distinguished by the human eyes and hardly be convincing. With the improvement of DeepFakes, the methods uses faked video inconsistencies in visual, audio or motion continuity to detect fake videos. Such as the LipForensics~\cite{Lips} apply high-level semantic irregularities in mouth movements, which are common in many generated videos.
The second category of DeepFake detection methods are data-driven~\cite{DeepfakesMTD,IDRevealDeepfake,MTDNet,RethinkingDeepfake}, not relying on any specific artifact.  Early researchers used some simple CNN structures. With the development of deep learning technology, advanced network structures such as Xception, LSTM and ViT have been applied to DeepFake detection and achieved better performance.
At present, the focus of researchers has shifted from improving the accuracy of identifying real/fake to providing reliable evidence \cite{Face_X-Ray,DeepfakeRepresentative}.

\subsection{Vision Transformer}
Transformer is proposed from natural language processing (NLP)~\cite{AttentionIsAll}. In the last few years, the ViT-based networks have shown great success in a wide range of domains including but not limited to image and video tasks \cite{ViViT,HRFormer,MetaFormer,jiang2021transgan,FuseFormerGan}.
Vision Transformer is an effective feature extraction structure for video, especially for sequence-to-sequence modeling \cite{AttentionIsAll,PVT,pvtv2}.  In the early study, the ViT-based models are known to only be effective when large training datasets are available. The parameters and calculated cost of ViT structure increase exponentially with the increase of image pixels (same  size). To solve this problem, researchers proposed many improved ViT architectures. Swin Transformer~\cite{liu2021Swin} shifted windowing scheme brings greater efficiency by limiting self-attention computation to non-overlapping local windows while also allowing for cross-window connection. This hierarchical architecture has the flexibility to model at various scales and has linear computational complexity with respect to image size. HRFormer~\cite{HRFormer} take advantage of the multi-resolution parallel design introduced in high-resolution convolutional networks, along with local-window self-attention that performs self-attention over small non-overlapping image windows, for improving the memory and computation efficiency. Recent studies have combined CNN and ViT and achieved better performance~\cite{MetaFormer,guo2021cmt}.

\subsection{Hashing Retrieval}
The hashing retrieval method maps the high-dimensional content features of the image or video to Hamming space (binary space), which decreases the requirements of the image or video retrieval system for computer memory space, improves the retrieval speed, and can better meet the requirements of huge data retrieval~\cite{Retrieval1,SplicedRetrieval,HashingRetrieval2,TBH,CSQ}.
Deep learning based hashing retrieval includes trained hash centers and pre-trained hash centers.
The trained hash centers maps the same classification data to a hash center, which aims to distinguish as many different samples as possible\cite{CSQ}.
The pre-generated hash centers generate an Near-Optimal hashes before training, which the Hamming distance between different hash centers is near half of the hash bits\cite{TBH}.
During training, similar samples are mapped to the same hash center.

\subsection{Tampering Localization}
Forgery detection is used to identify the real/fake of an image or video.
However, sometimes we also want to know the suspicious regions of forgery.
Tampering localization provides a pixel level classification~\cite{TL3,TL4,TL5}.
Image tampering localization can be categorized into image splicing, copy-move and object inpainting.
At present, researchers mainly focus on image tampering localization and pay less attention to video tampering localization.
Video tampering localization is mostly an extension of image tampering localization .
Early researchers used hand-crafted features to locate tampered images.
In recent years, CNN-based tampering localization methods have become mainstream~\cite{TL1,TL2,DMAC}.              
\section{Method}
\label{sec:method}
In this section, we first reorganize the structure of the video in the dataset to adapt the training in Sec.~\ref{subsec:data}.
Then, we use Hash Triplet Loss to learn hash centers in Sec.~\ref{subsec:learning-hash-centers}.
Finally, the trained hash centers and the model of ViTHash are saved to use to trace the source of fake videos in Sec.~\ref{subsec:tracing}.

\subsection{Data Preparation}
\label{subsec:data}
As shown in Fig.~\ref{fig:flowchart} Step1: Data preparation, we used the original video and related fake videos as a group.
When training, the original video and two fake videos as a unit.
Every unit contains the original videos, the batch size $n > 8$ ($3\times8$ videos each batch size) to ensure the uniform distribution of videos.
Therefore, the hash codes of all related fake videos will be as similar as possible to the original video.
In this way, ViTHash can adapt to more alternative fake videos, rather than only effective for several videos in the training set.

\subsection{Hash Centers Learning}
\label{subsec:learning-hash-centers}

\begin{algorithm}
\caption{The Whole Calculation Process of \textbf{Hash Triplet Loss}}
\label{alg:hashloss}
\SetKwInput{KwInput}{Input}                
\SetKwInput{KwOutput}{Output}              
\DontPrintSemicolon
  
  \KwInput{
    ViTHash outputted hashes and related labels as $HL_{s}$\;\hspace{.9cm} Voted Hash Centers and related labels as $HC_{s}$}
  \KwOutput{Loss of \textbf{Hash Triplet Loss} }

  \SetKwFunction{FMain}{Main}
  \SetKwFunction{FInterLoss}{InterLoss}
  \SetKwFunction{FIntraLoss}{IntraLoss}
  \BlankLine

\emph{Calculate the Intra-Loss between the triplet samples and the voted Hash Center}\;
  \SetKwProg{Fn}{Def}{:}{}
  \Fn{\FIntraLoss{$\vec{h}$, $\vec{HC_{i}}$}}{
    \KwRet $mean(\left | \vec{h}-\vec{HC_{i}}\right | )$\;
  }
  \;
  
\emph{Calculate the Inter-Loss between the triplet samples with the other Hash Centers}\;
  \SetKwProg{Fn}{Def}{:}{}
  \Fn{\FInterLoss{$\vec{h}$, $\vec{HC_{i}}$}}{
    \KwRet $1 - mean(\left | \vec{h}-\vec{HC_{i}}\right | )$\;
  }
  \;
  
\emph{Calculate the Hash Triplet Loss}\;
  \SetKwProg{Fn}{Function}{:}{\KwRet}
  \Fn{\FMain}{
    $inter,intra,n,m=0,0,0,0$\;
    \For{$label_{h}, \vec{h}$\textbf{ in }$HL_{s}$}{
      \For{$label_{o}, \vec{o}$\textbf{ in }$HC_{s}$}{
      \eIf{$label_t==label_v$}{
        $m+=1$\;
        $intra+ =IntraLoss(\vec{h},\vec{o})$ \;
      }{
        $n+=1$\;
        $inter+ =InterLoss(\vec{h},\vec{o})$ \;
      }
    }\;
    }\;
    \KwRet $intra \div m + intra \div n$\;
  }
  
\end{algorithm}

The whole calculation process of \textbf{Hash Triplet Loss} $\mathcal{L}$ as shown in Alg.~\ref{alg:hashloss}.
The main idea of $\mathcal{L}$ here is enlarging the \textbf{Inter-Class} loss and lessening the \textbf{Intra-Class} loss.
As shown in Algo.~\ref{alg:hashloss}, where $n$ is the number of the \textbf{Inter-Class} samples of different labels, $m$ is the number of the \textbf{Intra-Class} samples of the same labels.
The $\mathcal{L}$ is defined as
\begin{equation}
   \dis{\vec{x}}{\vec{y}}=\overline{\left | \vec{x}-\vec{y}   \right | }  
\end{equation}
\begin{equation}
\mathcal{L}= \sum\limits_{i=1}^{m}\dis{\vec{v_{i}}}{\vec{h_{i}}}/{m} +
    \sum\limits_{j=1}^{n}(1-\dis{\vec{v_{j}}}{\vec{h_{j}}})/n \label{eq:loss_h}
\end{equation}
ViTHash learning hash centers in supervised mode.
Given the training set $\mathbb{X} = \left \{{x_{i}}\right \}_{i=1}^{z}$, original videos as $\mathbb{S} = \left \{{s_{i}}\right \}_{i=1}^{m}$, fake videos as $\mathbb{F} = \left \{{f_{i}}\right \}_{i=1}^{n}$, and $z=m+n$.
We define the hash centers as $\mathbb{HC}_{s} = \left \{{HC_{i}}\right \}_{i=1}^{m}$, where m is the number of original videos.
ViTHash outputs a hash binary code denoted as $\vec{h}  \subset \{0,1\} ^{k}$, where $k$ is the hash bits.
We define the videos of each group which with the same hash center as $\mathbb{X}_{i}= \{s_{i},f_{i1},f_{i2},\dots,f_{ij} \} $.
The optimization purpose of $\mathbb{HC}_{s}$ is to find a $HC_{i}$ that minimizes the Hamming distance between $\mathbb{X}_{i}$ with $HC_{i}$, and the mean Hamming distance of $\mathbb{HC}_{s}$ is $\tfrac{1}{2} k$.

As shown in Fig.~\ref{fig:flowchart} Step2: Learning hash centers.
Inspired by K-means, we adopt an idea that gradually adjusts the hash centers by \textbf{Hash Triplet Loss}, see Alg.~\ref{alg:hashloss}.  \textbf{ViTHash} training $\mathbb{HC}_{s}$ by triple videos $\mathbb{U}_{i}= \{s_{i},f_{ij},f_{ik}\}$.
The $\mathbb{U}_{i}$ produces a temporary hash center $\vec{h}_{i}$ through voting algorithm.
The \textbf{Hash Triplet Loss} lessens the Hamming distance of hash binary code between $x \in \mathbb{X}_{i}$ and $\vec{h}_{i}$, and enlargement the distance between $\mathbb{X}_{i}$ and $\mathbb{X}_{j}$.
Then $\mathbb{HC}_{s}$ will be updated at each training iteration.
Finally, the optimized $\mathbb{HC}_{s}$ are outputted.

\subsection{Tracing the Source of Fake Videos}
\label{subsec:tracing}
As shown in Fig.~\ref{fig:flowchart} Step3: Tracing the source.
Tracing the source of fake videos is a simple task.
After training hash centers $\mathbb{HC}_{s}$ with ViTHash, we saved $\mathbb{HC}_{s}$ and the model of ViTHash.
A fake video denoted as $f_{i}$ to generate a hash code $\vec{h_{i}}$ by \textbf{ViTHash}.
Next, find the $HC_{i}$ in $HCs$ by calculating the minimum Hamming distance with $\vec{h_{i}}$, the $s_{i}$ is the original video of $HC_{i}$.
After that, compare $f_{i}$ with $s_{i}$.
\section{Networks}
\label{sec:networks}

\subsection{Overview of Networks}
\label{subsec:overview}
\begin{figure*}[t]
    \begin{center}
        \includegraphics[width=1\linewidth]{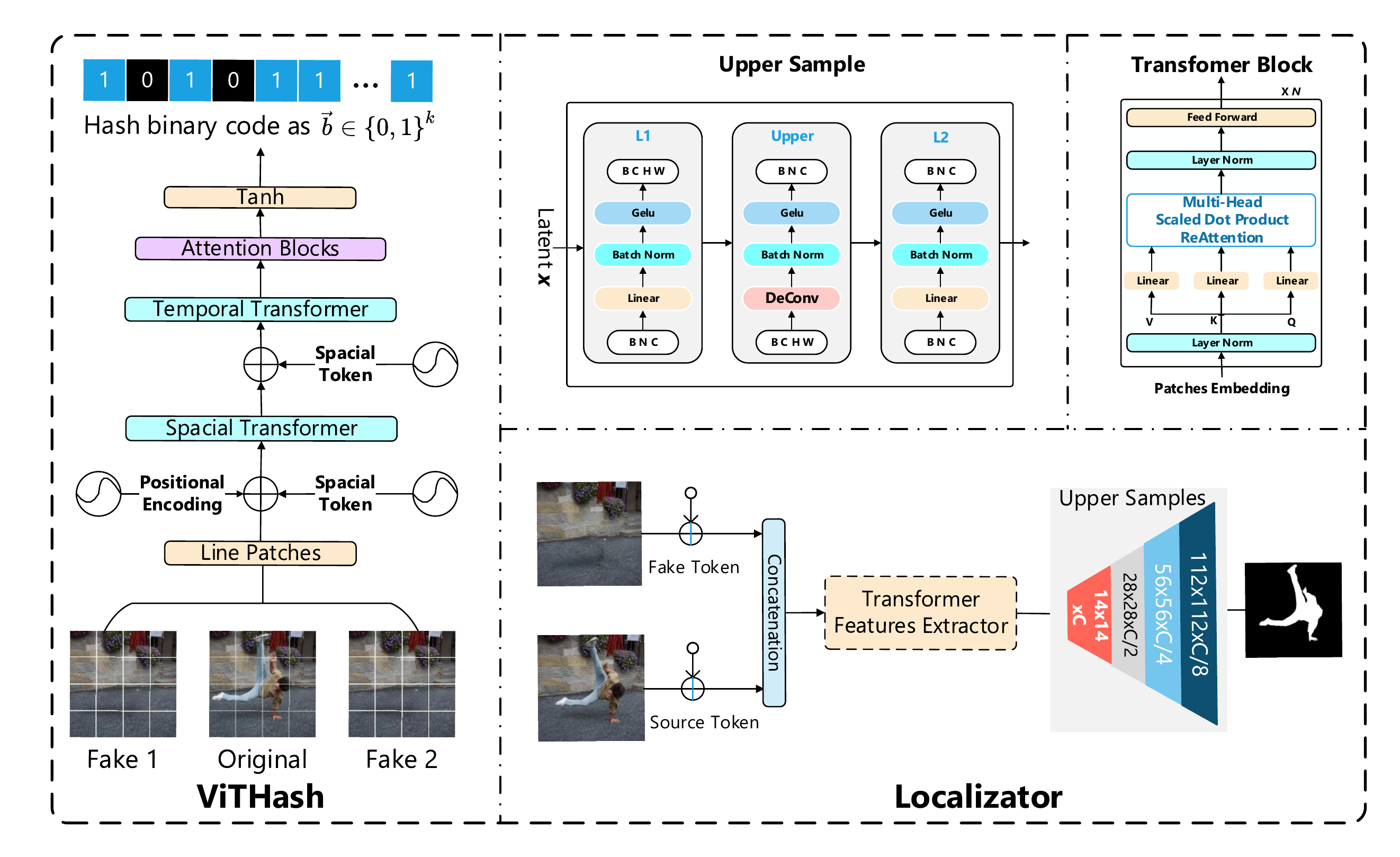}
    \end{center}
    \caption{
       Overview of our proposed networks. Our method contains two networks: \textbf{ViTHash} and \textbf{Localizator} , and two basic modules: \textbf{Upper Sample} and \textbf{Transformer Block}.
        \textbf{ViTHash} and \textbf{Localizator} consist of basic modules.
         \textbf {ViTHash} training hash centers from triplet videos: the original video and two randomly related fake videos. The trained hash centers are used to trace the source of fake videos.
        The \textbf{Localizator} we designed to analyze the differences between the traced video and the fake video, which is not affected by video quality and cropping. The different areas of the two videos are represented by generated masks.
    }
    \label{fig:arch}
\end{figure*}

\paragraph{ViTHash}
As shown in Fig.~\ref{fig:arch} \textbf{ViTHash}.
ViTHash is used to train hash centers and tracing the source.
The feature extraction of ViTHash is composed of series spatio-temporal Pyramid Vision Transformer (PVTv2)~\cite{pvtv2}, and multiple Attention Blocks.
The first module Transformer Block focuses on spatial features.
The second module Transformer Block  focuses on temporal features.
Finally, the results are output through the $\tanh$ function.
The output of $\tanh$ is converted into binary code by
\begin{equation}
        \sign{x} =
    	\left \{
        		\begin{aligned}
            		\; 1 & \quad & x \geq 0 \\
            		\; 0  & \quad & x < 0
    		\end{aligned}
    	\right.
\end{equation}
\begin{equation}
    \Vec{b}=(\sign{x}+1)/2 \in \left \{0,1 \right \}^{k}   \label{eq:bin}
\end{equation}
where $k$ is the hash bits.

\paragraph{Localizator}
As shown in Fig.~\ref{fig:arch} \textbf{Localizator}.
Localizator is a comparison tool for the traced video and the fake video in this paper.
It is convenient for users to intuitively see the difference between the traced result and the fake video, especially when there are many tasks.
We observed the ViT-based networks that destroy the space continuity of pixels when training with linear patches images.
To improve performance, we designed a CNN-ViT mixed structure.
The CNN blocks are used to learn high-level features and focus on the correlation of local pixels. The ViT focuses on the long-range context and temporal dimension features of video.
Besides, we design an Upper Sample to learn more details for discrepancy regions.

\subsection{Why do we choose the hashing method?}
\label{subsec:choice}

\paragraph{Fast and Low Cost}
We assume that different methods of backbone networks, the time cost needed for once detection of the model is not much different, which is $t$.
The time cost of the traditional forgery detection method is $t_{1}=t$.
The time cost of the hashing retrieval method is $t_{2}=\lambda +t \approx t$, where $\lambda$ is the time cost of calculating the Hamming distance.
The time cost of content matching retrieval method is $t_{3}=n \times t$, where $n$ is the number of matching videos.
Hashing retrieval also takes up tiny space to store the hash code and index of the video.
The hash code is a fixed length binary bit (k bits), and an index is represented by a 32-bit integer.
The total storage space occupied is $(32+k)\times n$, where $n$ is the number of original videos.

\paragraph{Better Versatility}
The ViTHash does not depend on the semantics and forgery methods of videos.
We have done extensive experiments on public datasets for Deepfake.
At the same time, we have built several datasets to verify the performance of object inpainting and video splicing scenes.
These experiments shown that our method has superior performance for different forged videos.
We also published the datasets on the internet to ensure the credibility and fairness of our experiments.
The details of datasets and experiments are described in Sec.~\ref{sec:Experiments}.

\paragraph{Reliability}
Traditional forgery detection method can not provide completely reliable results, including those that claim to provide additional interpretable visual features.
ViTHash finds the related original video of fake videos by tracing the source.
The result of comparing the original video with the forged video is irrefutable.
Actually, the difficulty of this paper is how to improve the accuracy of retrieval.

\section{Experiments}
\label{sec:Experiments}
\begin{figure*}[ht]
    \centering
    \subfloat[DFTL]
    {
        \label{fig:db_DFTL}
        \includegraphics[width=.23\linewidth,height=.23\linewidth]{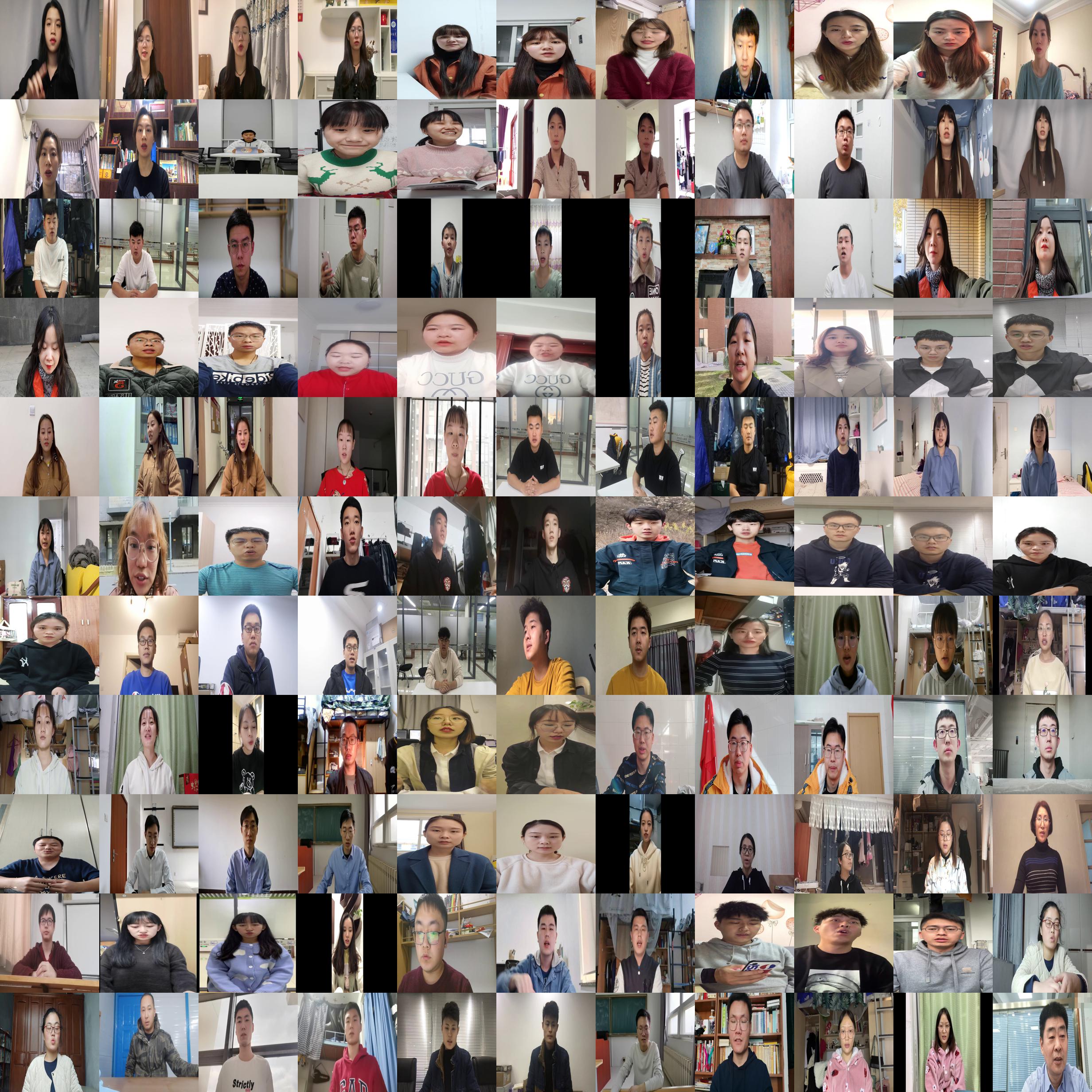}
    }
    \subfloat[VSTL]
    {
        \label{fig:db_VSTL}
        \includegraphics[width=.25\linewidth,height=.23\linewidth]{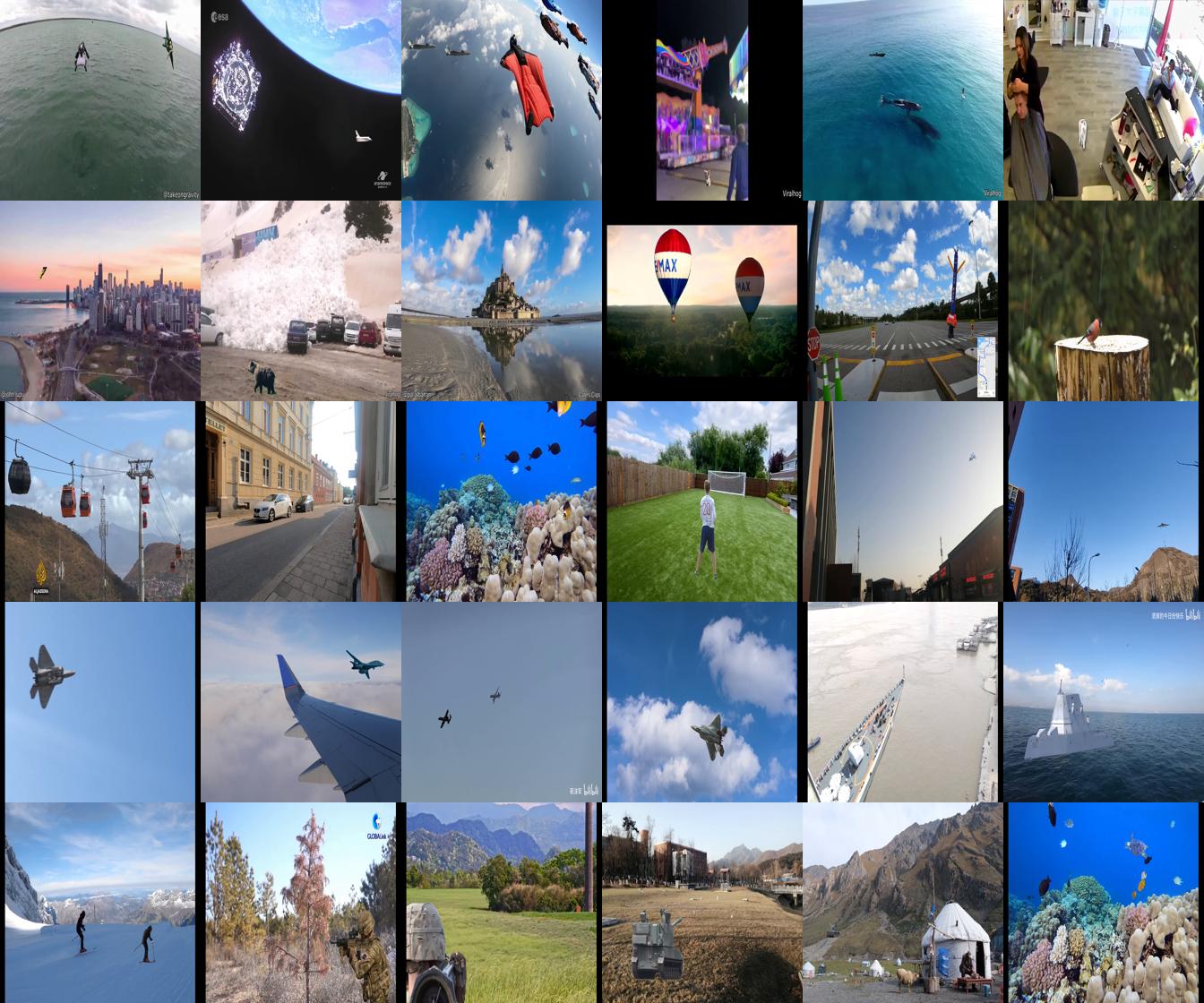}
    }
    \subfloat[DAVIS2016-TL]
    {
        \label{fig:db_Davis2016}
        \includegraphics[width=.23\linewidth,height=.23\linewidth]{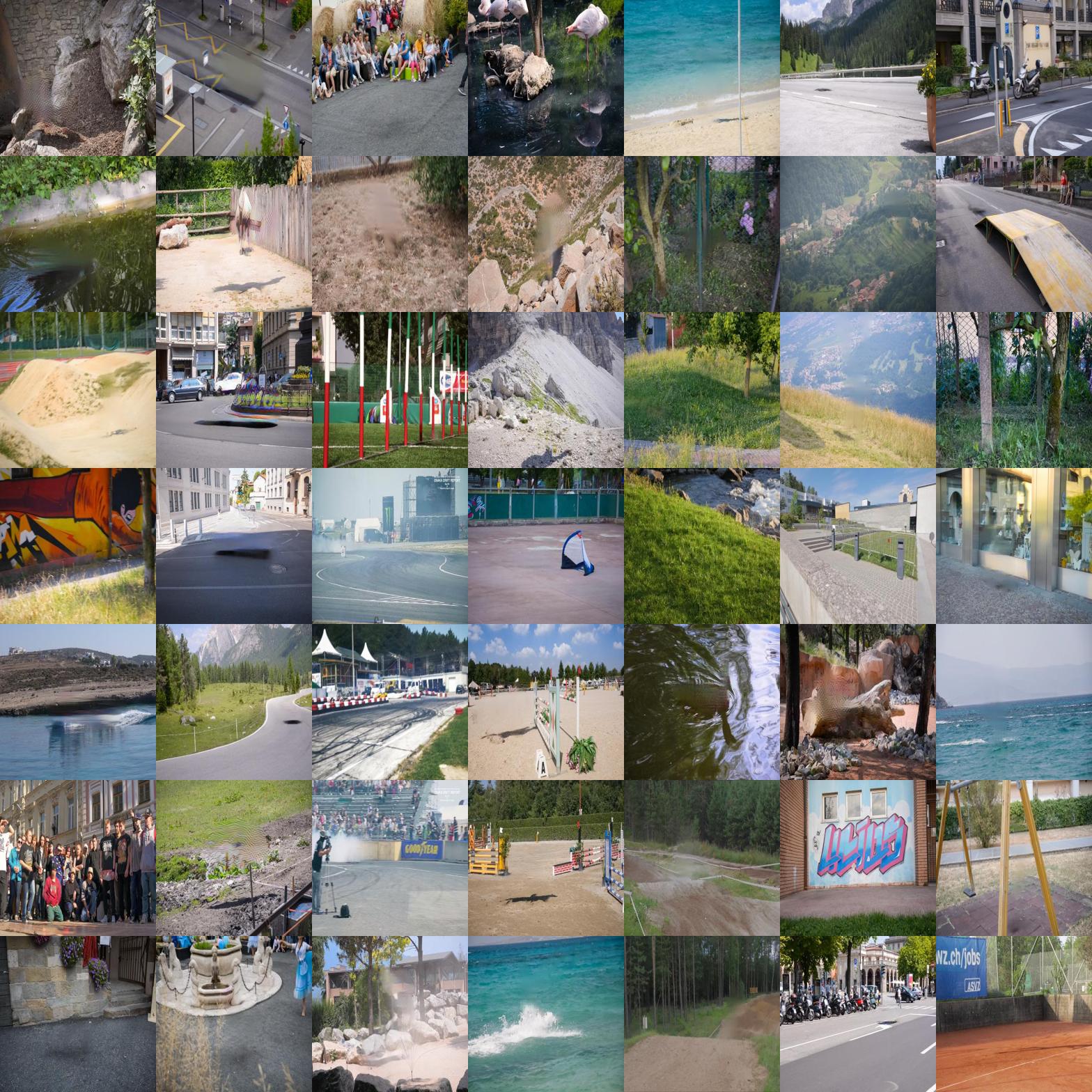}
    }\\
    \caption{
        (a) DFTL:We paid for 75 Asians and each one shot three videos for 15 minutes. We used three Deepfake methods to make fake videos and finally selected some high-quality videos.
        (b) VSTL: We have carefully manufactured 30 video object splicing videos, many of which are military scenes. The foreground and objects of the video are added frame by frame, and the added object size and position is reasonable.
        (c) DAVIS2016-TL: DAVIS2016-TL is based on the DAVIS2016 dataset and related masks. Using five state-of-the-art objects inpainting methods to synthetic 300 fake videos.
    }
    \label{fig:datasets}
\end{figure*}
\begin{figure*}[ht]
    \centering
    \subfloat[Similar scenes]
    {
        \label{fig:Similar-a}
        \includegraphics[width=.29\textwidth]{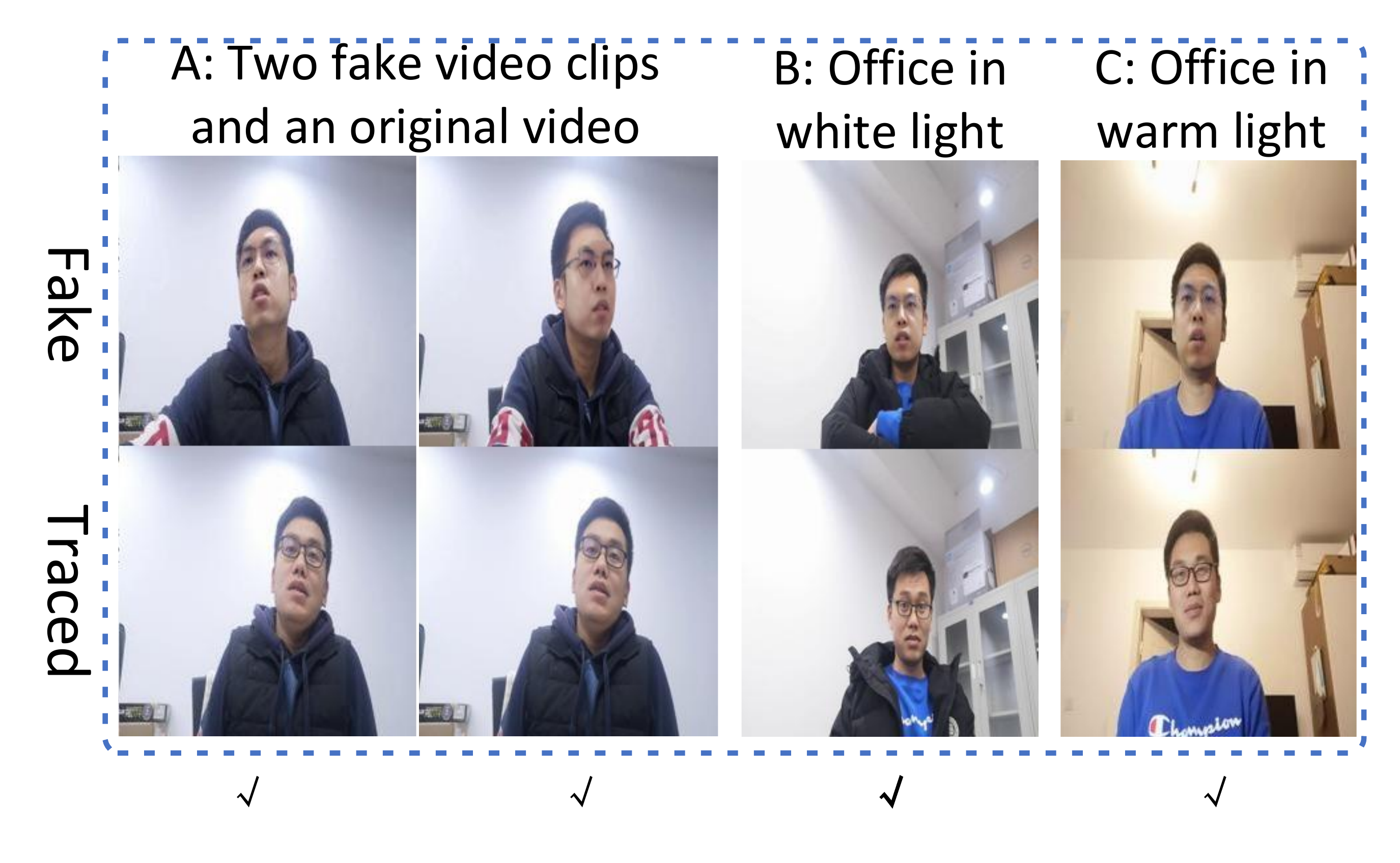}
    }
    \subfloat[Same scenes]
    {
        \label{fig:Similar-b}
        \includegraphics[width=.29\textwidth]{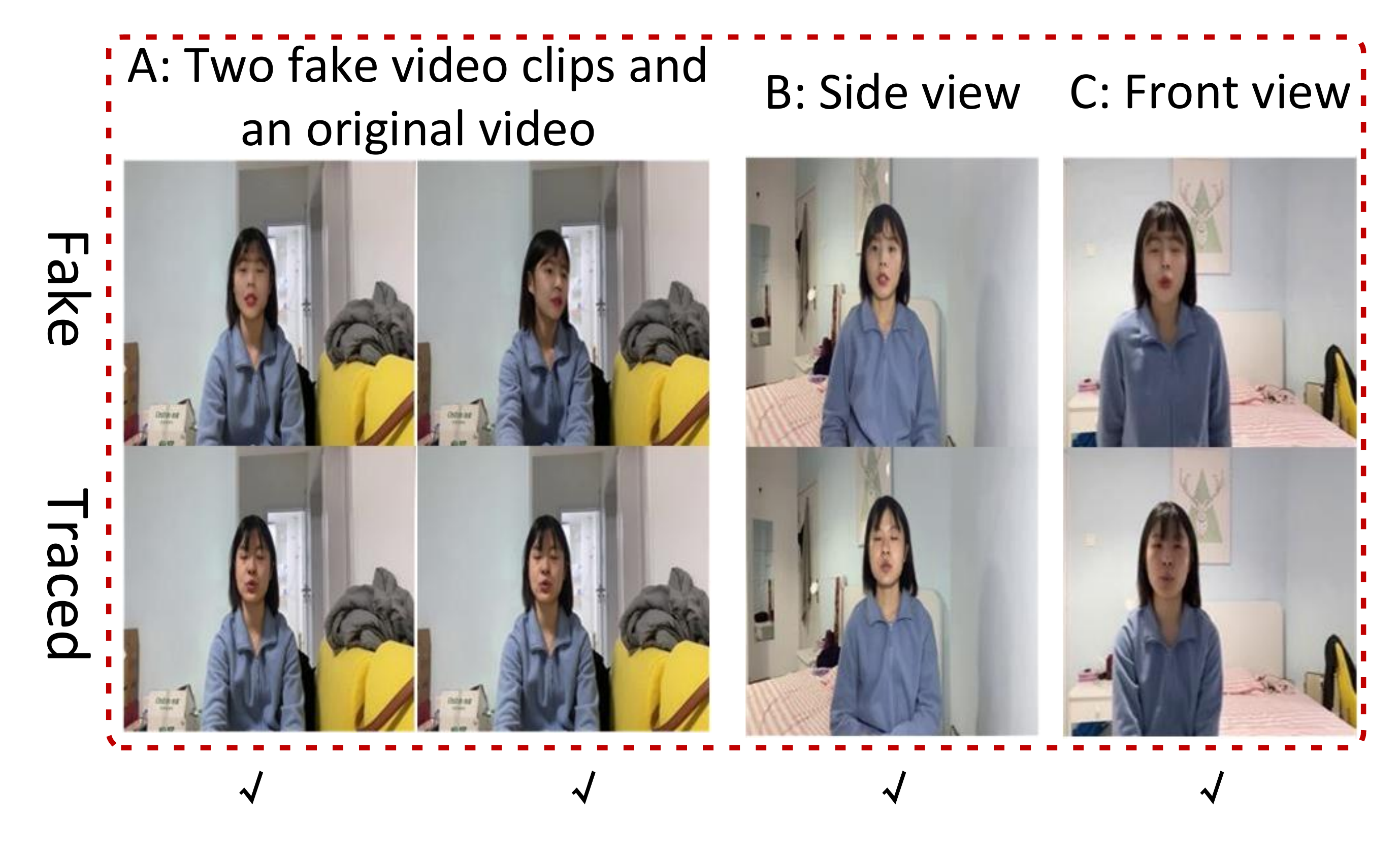}
    }
    \subfloat[Error Analysis]
    {
        \label{fig:error}
        \includegraphics[width=.34\textwidth]{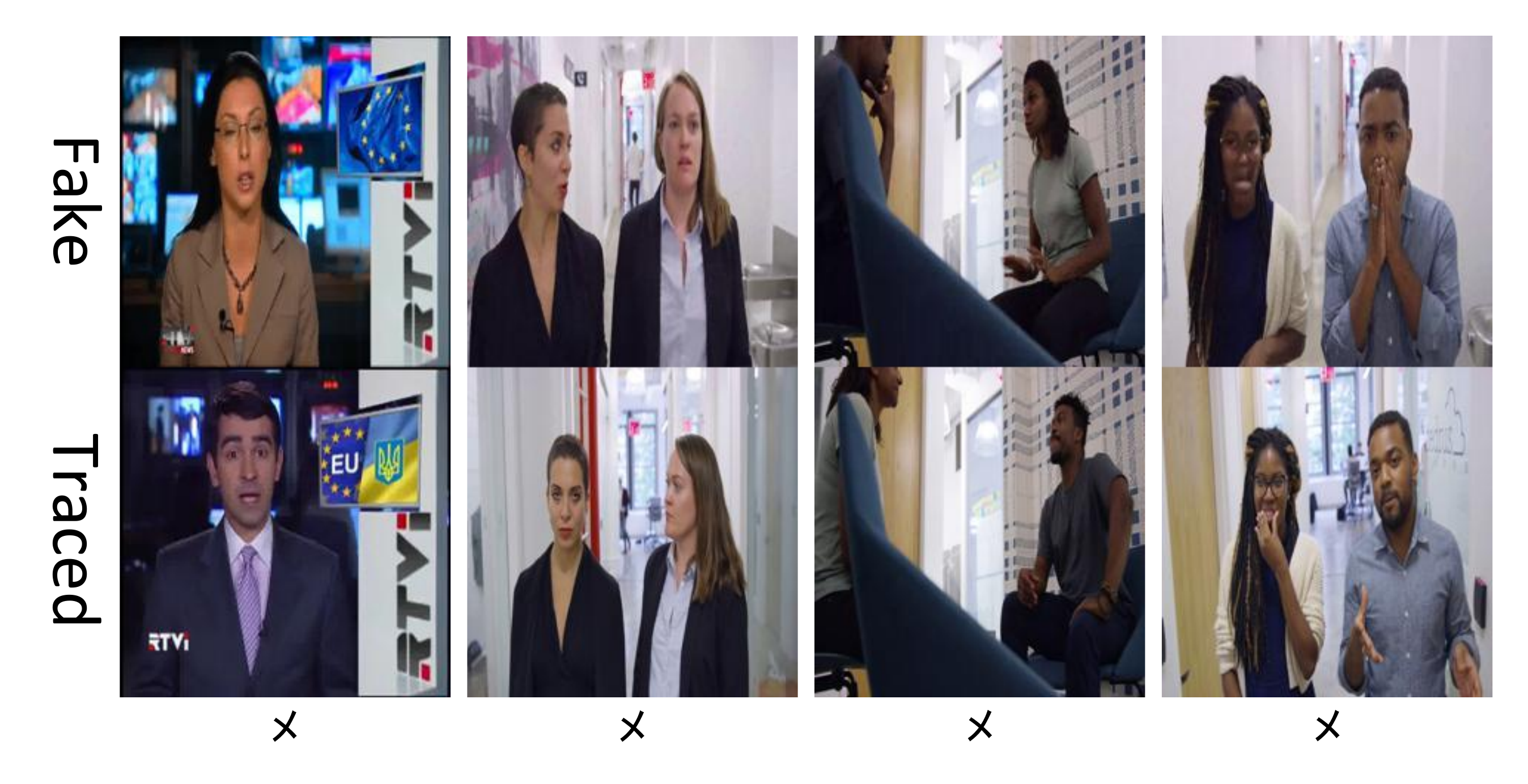}
    }\\
    \subfloat[FF++]
    {
        \includegraphics[width=.205\textwidth]{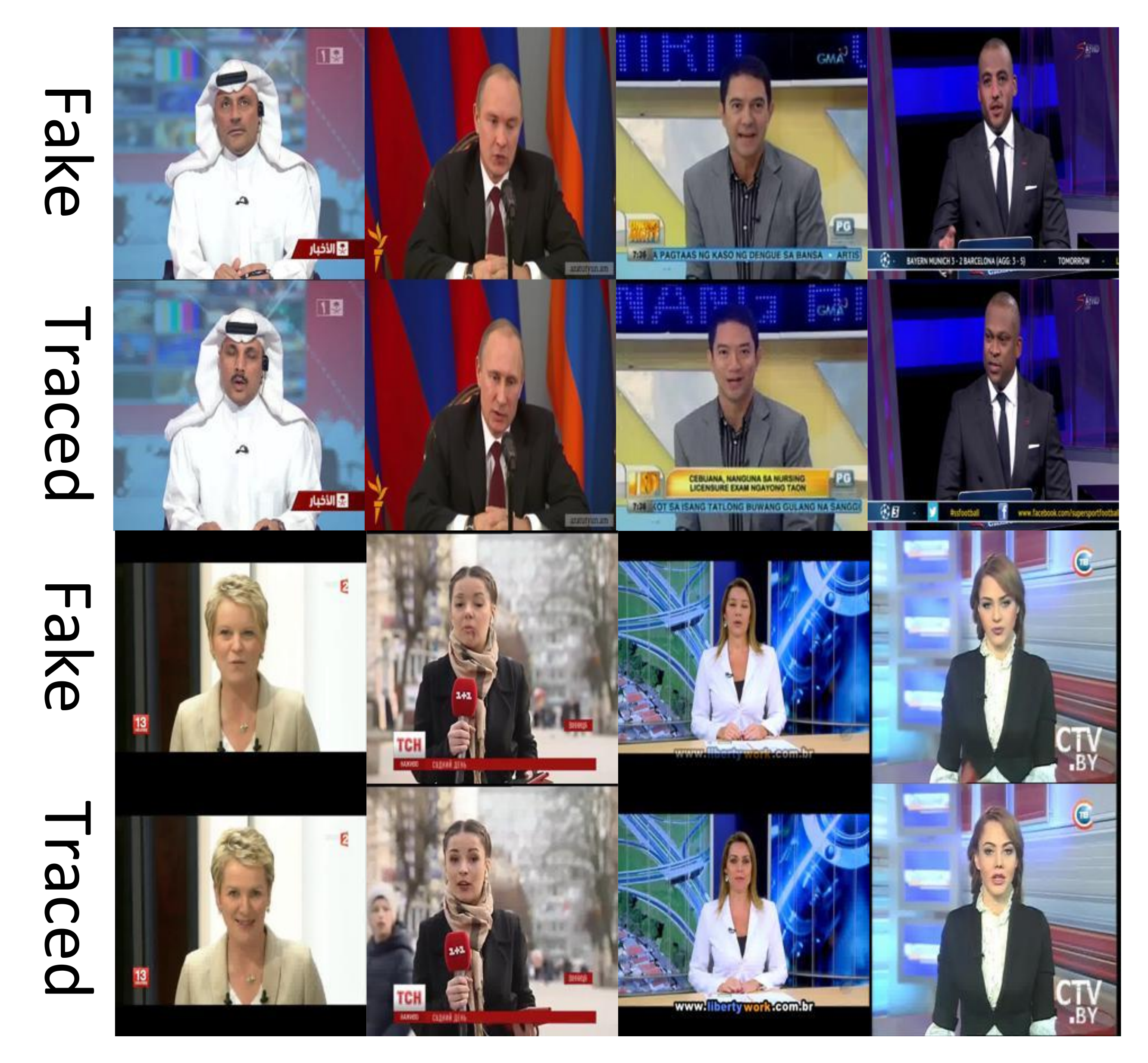}
    }
    \subfloat[DFD]
    {
        \includegraphics[width=.205\textwidth]{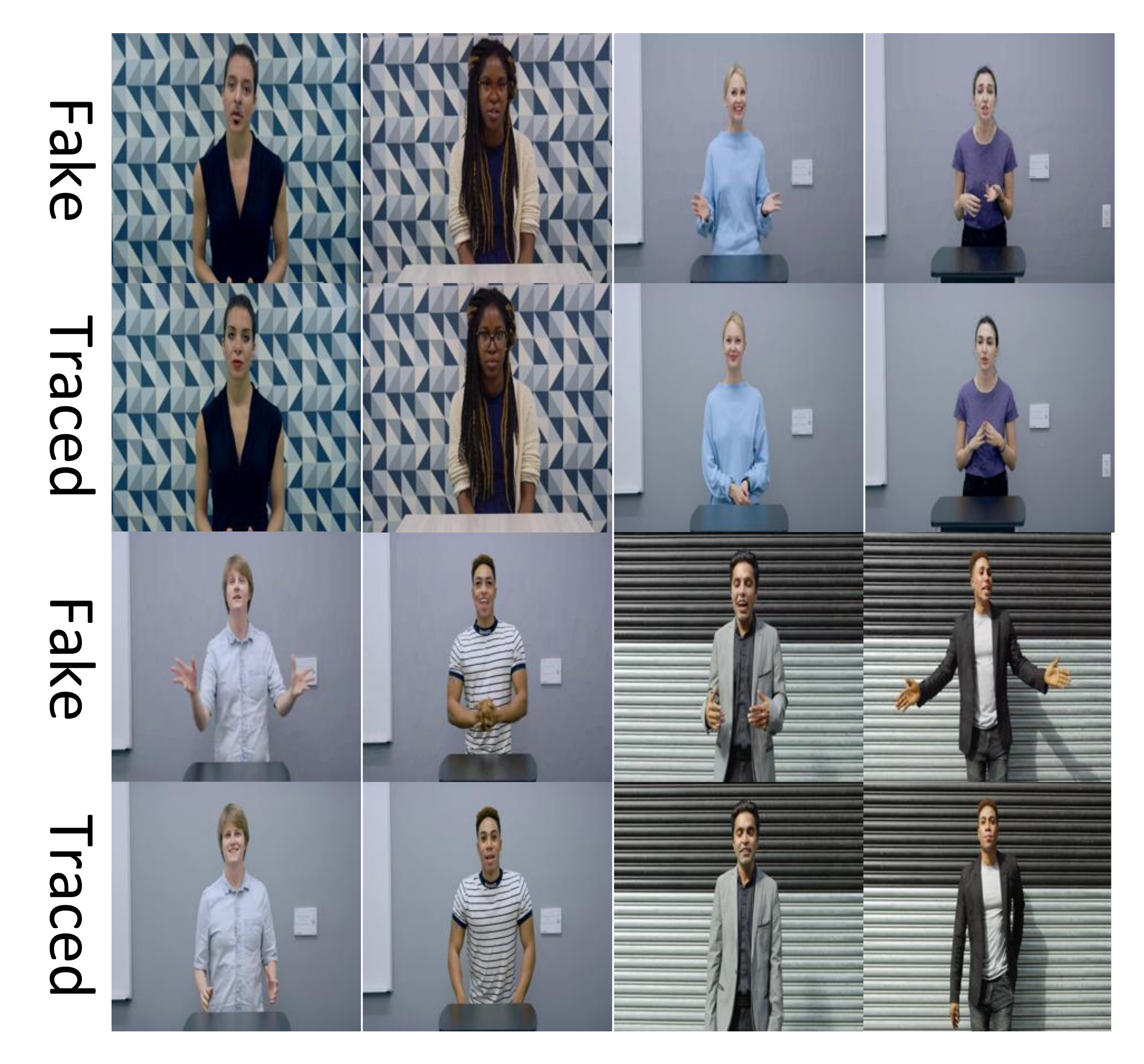}
    }
    \subfloat[VSTL]
    {
        \includegraphics[width=.205\textwidth]{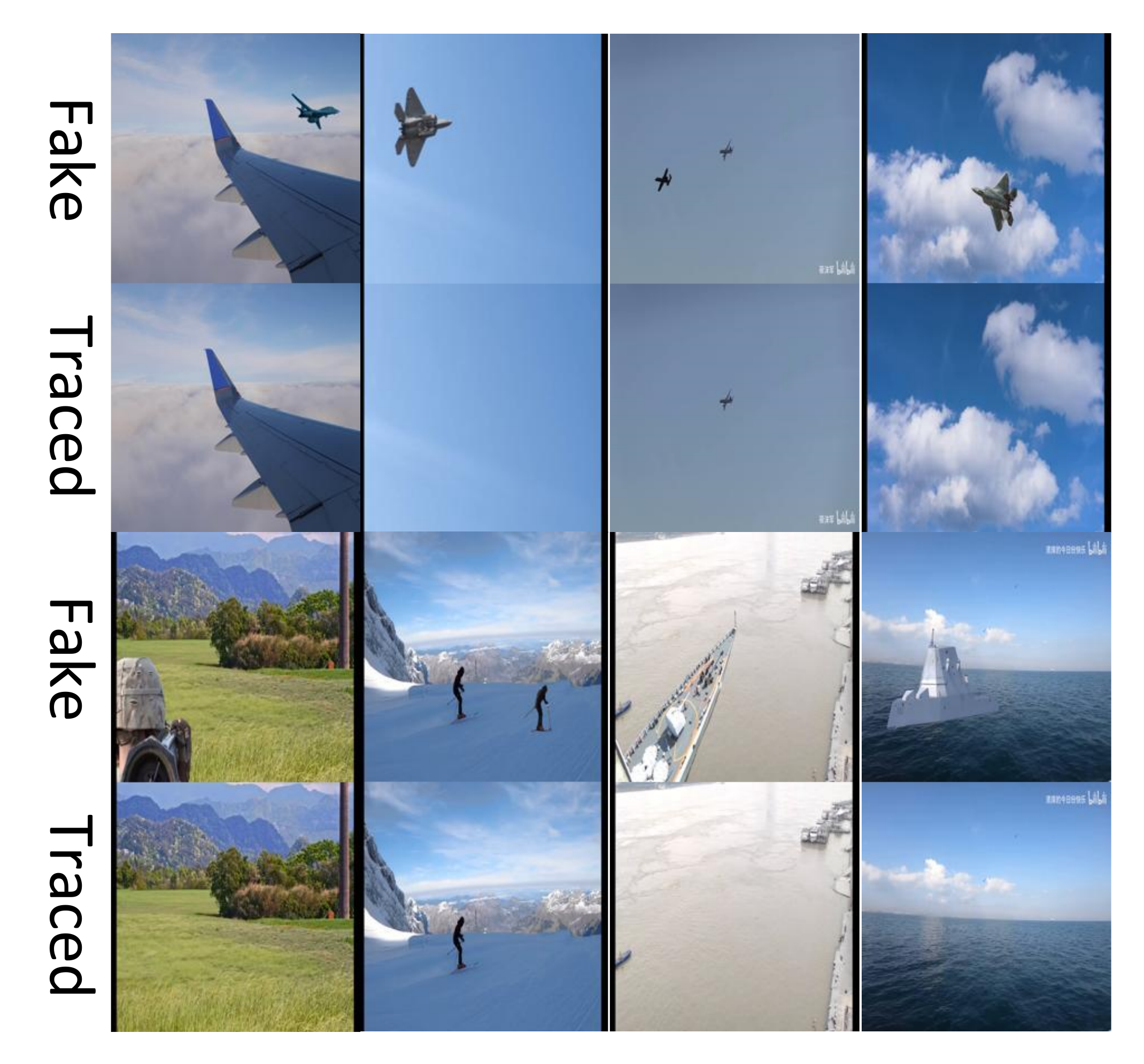}
    }
    \subfloat[DAVIS2016-TL]
    {
        \includegraphics[width=.205\textwidth]{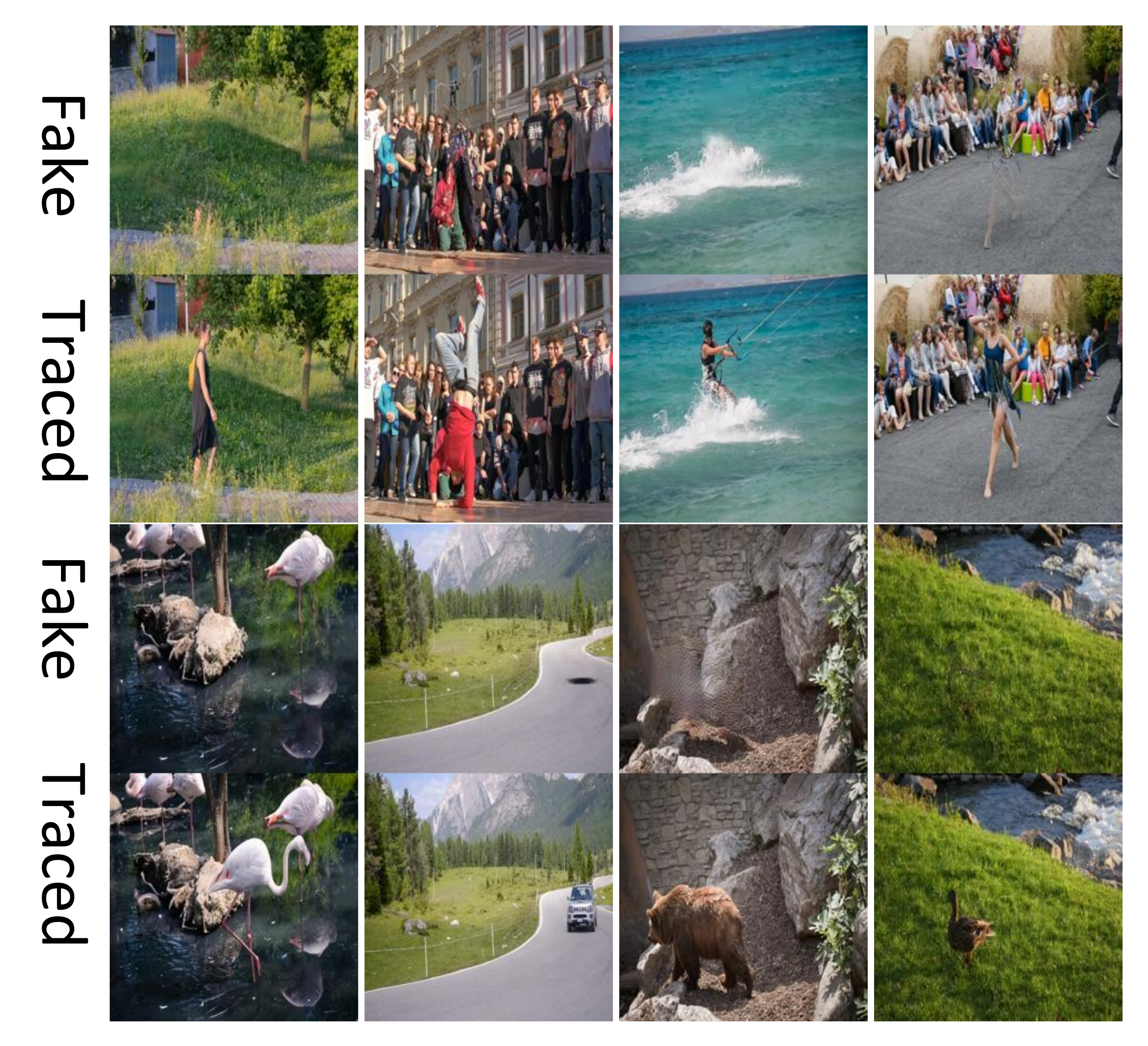}
    }\\
    \subfloat[VSTL]
    {
        \label{fig:l_a}
        \includegraphics[width=.205\textwidth,height=.205\textwidth]{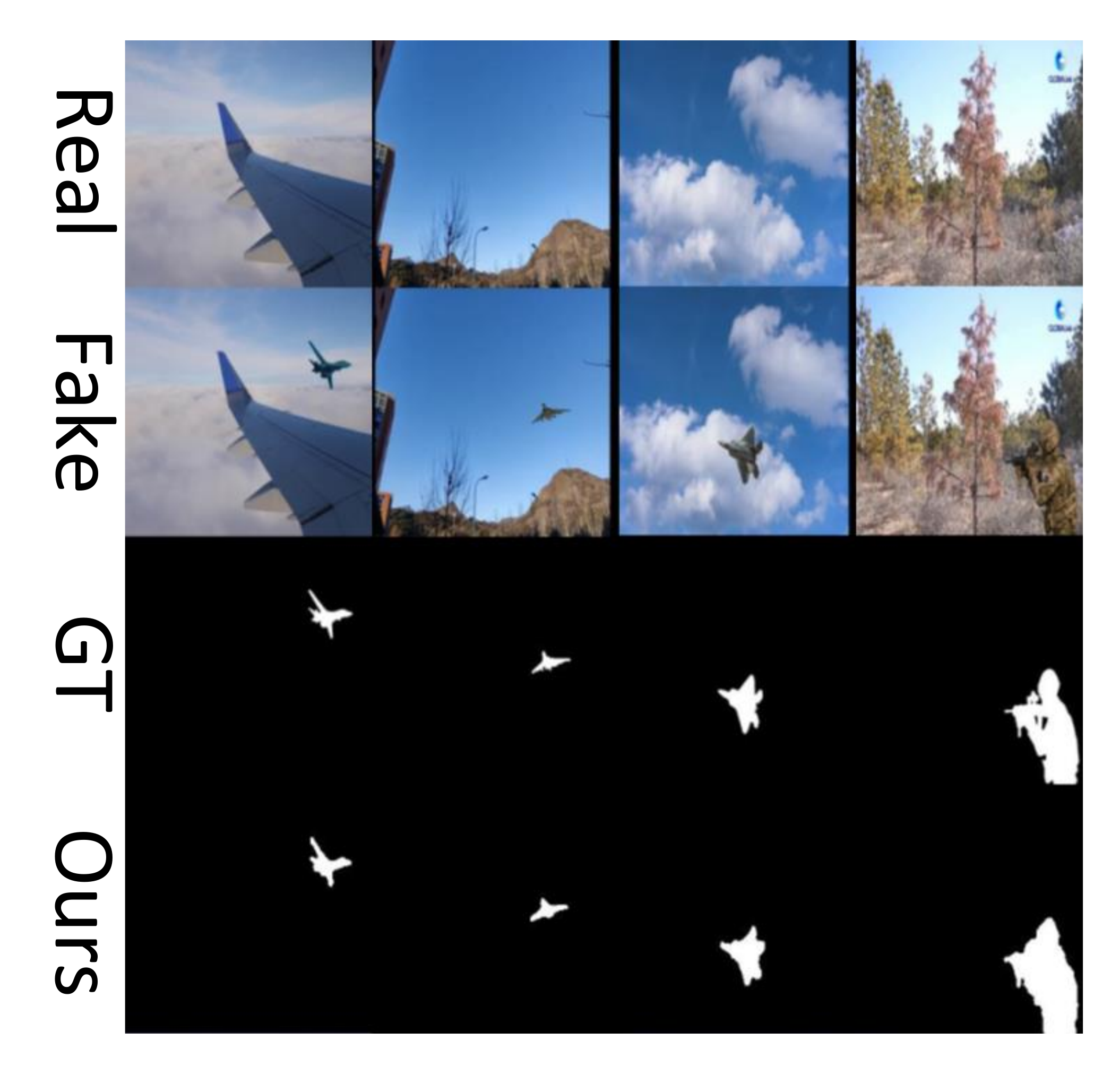}
    }
    \subfloat[DAVIS2016-TL]
    {
        \label{fig:l_b}
        \includegraphics[width=.205\textwidth,height=.205\textwidth]{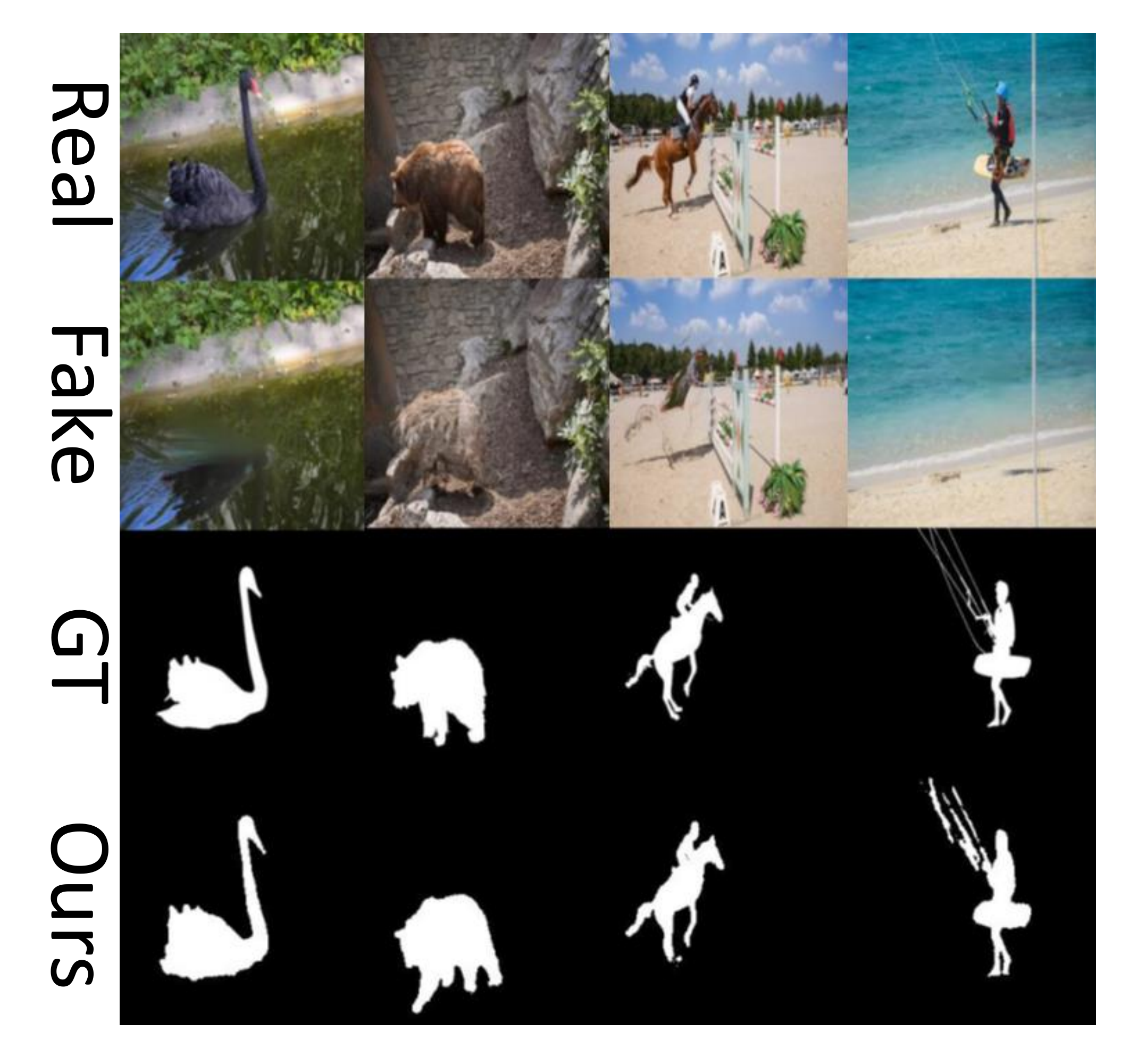}
    }
    \subfloat[DFD]
    {
        \label{fig:l_c}
        \includegraphics[width=.205\textwidth,height=.205\textwidth]{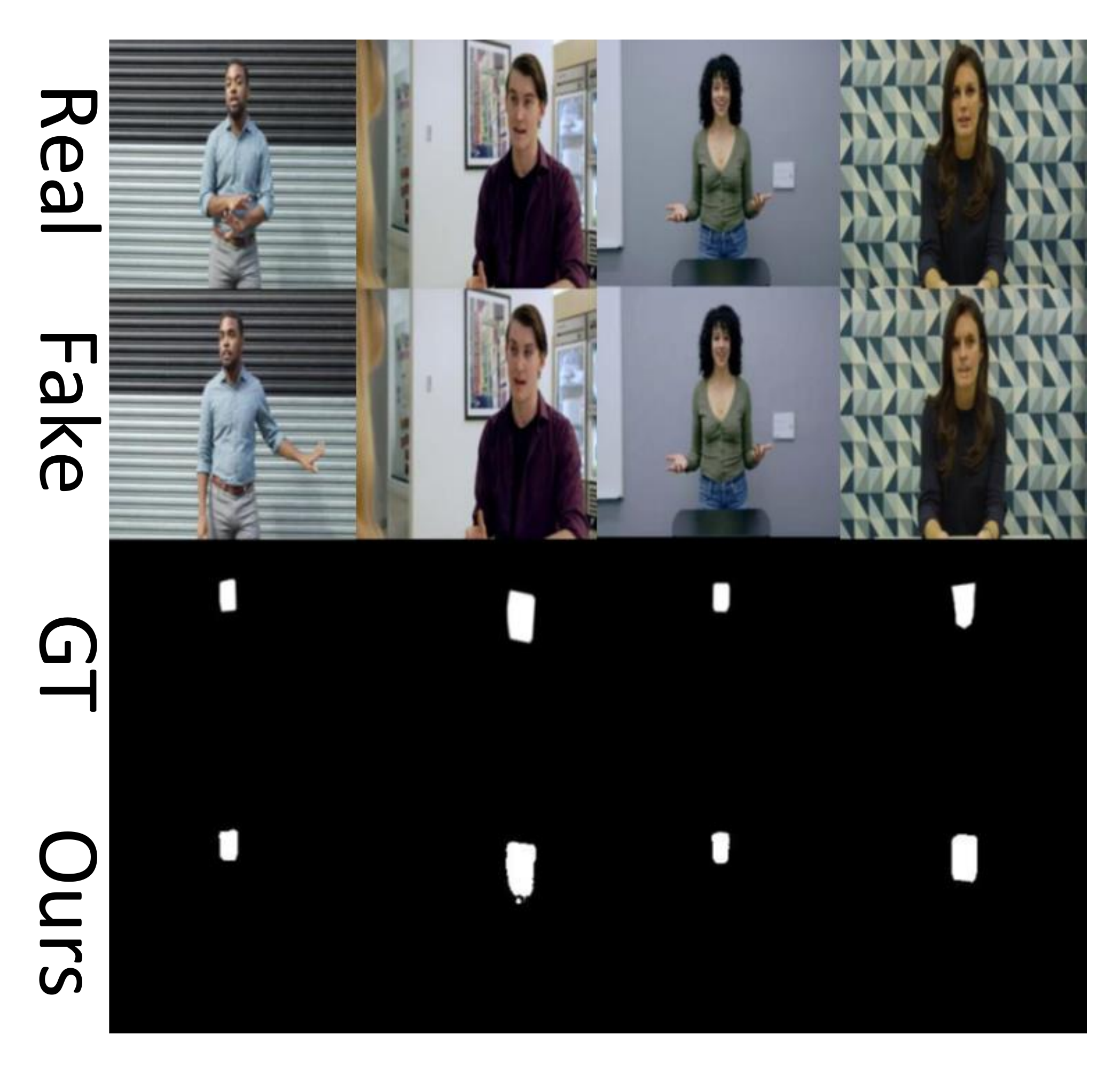}
    }\\
    \caption{
        (a) These videos are taken by the same person in a similar background. These scenes have different lights and similar backgrounds.
        (b) These videos are shot by the same person in the same scene. These videos are taken from different angles, and with the same clothes and postures.
        (c) These videos have very similar or the same background, only with different face areas.
        (d)-(g) The tracing results on four datasets: FaceForensics++, DFDC, VSTL and DAVIS2016-TL. Where the “traced” video is the original video in the dataset.
        (h)-(j) Tampering localization of comparing the fake videos with the original videos on three different datasets. Where the “GT” means “ground truth”.
    }
    \label{fig:Localization}
\end{figure*}

In this section, we design two different parts of the experiment.
One part is the evaluation of Deepfake, the other part is the evaluation of object-based forgery.
First, we introduce three public datasets and our built datasets in Sec.~\ref{subsec:dataset}.
Then, we describe the experiments setup and comparison methods of two parts in Sec.~\ref{subsec:setup}
We evaluate ViTHash with those datasets in Sec.~\ref{subsec:compare} and Sec.~\ref{subsec:measurement}.
Finally, we analyze the key issues in this paper in Sec.~\ref{subsec:analysis}, and discuss some necessary ablation experiments in Sec.~\ref{subsec:study}.

\subsection{Datasets}
\label{subsec:dataset}
\paragraph{Public Datasets}
We evaluate ViTHash on several public datasets of Deepfake.
The details of datasets are described as follows:
\begin{itemize}
    \item
    \textbf{FaceForensics++}.
    The FaceForensics++~\cite{roessler2019faceforensicspp, FF-dataset} dataset contains 1000 real videos from YouTube with over 500,000 frames. As well as the same number of manipulated videos generated by various of the state-of-the-art Deepfake methods. Moreover, the ratio of tampered videos and original videos is 1:1.

    \item\textbf{DeepFakeDetection}.
    The Google/Jigsaw Deepfake detection (DeepFakeDetection) dataset has~\cite{DFD} over 363 original videos from 28 consented actors of various genders, ages and ethnic groups. The DeepFakeDetection dataset contains 3,068 fake videos generated by four basic Deepfake methods.
    The ratio of original videos and fake videos is 0.12:1.

    \item\textbf{Celeb-DF}.
    The Full Facebook DeepFake detection challenge~\cite{Celeb-DF, Celeb-dataset} dataset is part of the DeepFake detection challenge, composed of 590 real videos and 5,639 fake videos. The ratio of synthetic clips and real clips is 1:0.23.
\end{itemize}

\paragraph{Our Built Datasets}
Due to the lack of related video datasets, we built three datasets and published them on the Internet to verify the reliability of ViTHash for various forgery methods.
These data sets may be limited and small, but they are sufficient to validate the metrics in this paper:
\begin{itemize}
    \item \textbf{DFTL}.
    We build a Deepfake Tracing and Localization (DFTL) dataset, and to verify Hash Triplet Loss tracing performance of similar videos.
    As shown in Fig.~\ref{fig:db_DFTL}, all actors are paid to use and modify their likeness.
    We use three methods: DeepFaceLab~\cite{DeepFaceLab,DeepFaceLab-Tool}, Faceswap~\cite{faceswap,fast-faceswap} and Faceswap-GAN~\cite{faceswap-GAN} to generate fake videos.
    Finally, we have chosen 187 videos (taken 1/10 frames of videos) from 75 people, including 133 training fake videos and 54 test fake videos, 578,613 frames in total.
    \item \textbf{DAVIS2016-TL}.
    As shown in Fig.~\ref{fig:db_Davis2016}, we've expanded the DAVIS2016\cite{DAVIS2016} dataset as DAVIS2016-TL, which is used to verify the performance of video object inpainting.
    DAVIS2016\cite{DAVIS2016} has 50 videos and object masks, and each video has 30--100 frames.
    Six methods are used to generate fake videos, including: FGVC~\cite{DBLP:conf/eccv/GaoSHK20}, DFGVI~\cite{DBLP:conf/cvpr/XuLZL19}, STTN~\cite{DBLP:conf/eccv/ZengFC20}, OPN~\cite{DBLP:conf/iccv/OhLLK19}, CPNET ~\cite{DBLP:conf/iccv/LeeOWK19} and DVI ~\cite{DBLP:conf/cvpr/KimWLK19a}. Finally, we obtained 200 training fake videos and 100 test fake videos, 33550 frames in total.
    \item \textbf{VSTL}.
    As shown in Fig.~\ref{fig:db_VSTL}, we also evaluate ViTHash on our built Video Splicing Tracing and Localization (VSTL) dataset.
    We made 30 videos of different scenes, including some military videos.
    The foreground of these videos are carefully selected to ensure that the spliced video objects can be properly integrated into the original video.
    These videos are made by a Photoshop like tool.
The frames of the original video as $\mathbb{A} =\left \{ a_{1}, a_{2},\dots,a_{n}\right \} $.
The frames of the spliced object as $\mathbb{B} =\left \{ b_{1}, b_{2},\dots,b_{m}\right \}$. The fake video as
\begin{align}
    \mathbb{R} & = \mathbb{A}+(scale\times\mathbb{B}+ pos) \\
    \mathbb{R}& = \left \{ r_{1}, r_{2} \dots r_{m}\right \}
\end{align}
Where $m < n$, $scale$ is scaling factor and $pos$ is the position $\mathbb{B}$ in $\mathbb{R}$.
    Because of limited resources, we randomly generated some spliced videos as training set.
\end{itemize}

\subsection{Experiments Setup}
\label{subsec:setup}

\paragraph{Implementation}
Our models are implemented by PyTorch, and the code has been released to GitHub.
We used ffmpeg to segment the video into frames, and training model with a single NVIDIA RTX 3090 24GB GPU card. Each model of the dataset trains 2-5 epochs.
Additionally, we use the Adaptive Moment Estimation (ADAM) optimizer with a learning rate of 1e-5. Being computationally efficient, ADAM requires less memory and outperforms on large datasets.

\paragraph{Comparison Methods of Deepfake}
We evaluated ViTHash on three public datasets: FaceForensics++, DeepFakeDetection and Celeb-DF, and compared ViTHash with three state-of-the-art methods:
\begin{enumerate}
\item 
Xception~\cite{Xception} based Deepfake detection method. Xception proposes a novel deep convolutional neural network architecture inspired by Inception, where Inception modules have been replaced with depthwise separable convolutions.
\item 
Grad-CAM~\cite{Grad-CAM} highlights the manipulated region with a supervised attention mechanism and introduces a two-stream structure to exploit both face image and facial detail together as a multi-modality task. Grad-CAM disentangling the face image into 3D shape, common texture, identity texture, ambient light, and direct light as a clue to detect subtle forgery patterns.
\item 
The face X-ray~\cite{Face_X-Ray} based on a common step: blending the altered face into an existing background image. So face X-ray provides a more general forgery detection , also apply to unseen face manipulation techniques The face X-ray of an input face image is a greyscale image that reveals whether the input image can be decomposed into the blending of two images from different sources.
\end{enumerate}

\paragraph{Comparison Methods of Object-Based Forgery}
We have done some additional experiments to verify the reliability of ViTHash on three our built datasets.
We also designed a tool to compare the fake video and the traced original video named Localizator.
The main purpose of the tool is to help us quickly localize the difference between the fake video and the original video.
Then we compare Localizator with two related methods DMAC~\cite{DMAC} and DMVN~\cite{DMVN} on VSTL and DAVIS2016-TL:
\begin{enumerate}
\item 
DMAC~\cite{DMAC} uses atrous convolution to adopt extract features with rich spatial information, the correlation layer based on the skip architecture is proposed to capture hierarchical features, and atrous spatial pyramid pooling is constructed to localize tampered regions at multiple scales.
\item 
DMVN~\cite{DMVN} estimates the probability that the donor image has been used to splice the query image, and obtain the splicing masks for both the query and donor images.
\end{enumerate}

\subsection{Evaluations of DeepFake}
\label{subsec:compare}

\begin{table*}
\centering
\caption{Deepfake videos detection results of tracing evaluation on FaceForensics++}
\label{tbl:hashbits}
\begin{tabular}{l|l|l|l|l|l|l|l|l|l|l|l|l|l|l|l} 
\toprule
\multicolumn{1}{c|}{\multirow{2}{*}{\begin{tabular}[c]{@{}c@{}}\textbf{Data}\\\textbf{Manipulation}\end{tabular}}} & \multicolumn{5}{c|}{\textbf{FF++ Raw (bits)}}                                      & \multicolumn{5}{c|}{\textbf{\textbf{FF++ C23 (bits)}}}                             & \multicolumn{5}{c}{\textbf{\textbf{\textbf{\textbf{FF++ C40 (bits)}}}}}             \\ 
\cline{2-16}
\multicolumn{1}{c|}{}                                                                                              & 64             & 128            & 256            & 512            & 1024           & 64             & 128            & 256            & 512            & 1024           & 64             & 128            & 256            & 512            & 1024            \\ 
\hline
\textbf{Original}                                                                                                  & \textbf{0.852} & 0.932          & 0.948          & 0.998          & \textbf{0.991} & \textbf{0.847} & 0.944          & 0.944          & 0.998          & 0.990          & 0.846          & 0.941          & \textbf{0.946} & 0.997          & 0.991           \\
\textbf{Detail}                                                                                                    & 0.850          & 0.930          & \textbf{0.949} & \textbf{0.999} & \textbf{0.991} & 0.845          & 0.943          & \textbf{0.945} & \textbf{0.999} & 0.989          & 0.847          & 0.942          & 0.945          & 0.996          & 0.991           \\
\textbf{Gaussian Blur}                                                                                             & 0.844          & \textbf{0.937} & 0.944          & \textbf{0.999} & 0.990          & 0.844          & 0.933          & 0.940          & \textbf{0.999} & \textbf{0.991} & \textbf{0.853} & \textbf{0.944} & 0.942          & \textbf{0.999} & 0.991           \\
\textbf{Blur}                                                                                                      & 0.846          & 0.934          & 0.947          & 0.998          & \textbf{0.991} & 0.844          & 0.944          & 0.939          & \textbf{0.999} & \textbf{0.991} & 0.848          & 0.942          & 0.941          & \textbf{0.999} & 0.991           \\
\textbf{Median Filter}                                                                                             & 0.850          & 0.935          & 0.948          & 0.998          & \textbf{0.991} & 0.844          & \textbf{0.945} & 0.941          & 0.998          & \textbf{0.991} & 0.851          & 0.942          & \textbf{0.946} & 0.997          & \textbf{0.992}  \\
\textit{\textbf{Cropping}}                                                                                         & 0.633          & 0.801          & 0.862          & \textit{0.983} & 0.963          & 0.636          & 0.862          & 0.814          & 0.986          & 0.962          & 0.629          & 0.816          & 0.859          & 0.988          & 0.964           \\
\bottomrule
\end{tabular}
\end{table*}

\begin{figure*}[htbp]
    \begin{center}
    \subfloat[Accuracy of robustness]
    {
        \label{chart:data}
        \includegraphics[width=.23\textwidth]{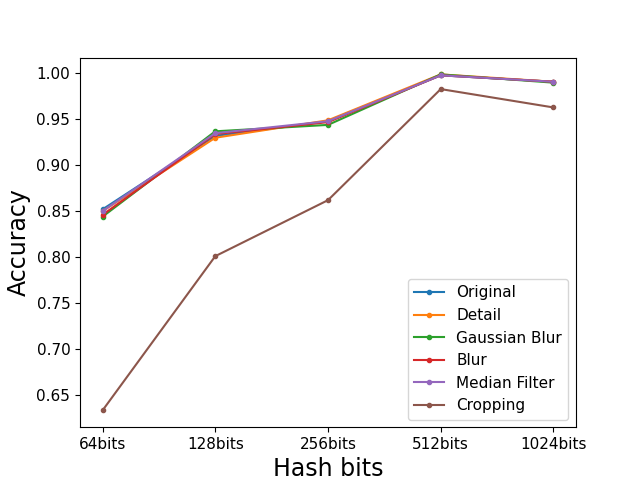}
    }
    \subfloat[Accuracy of compression]
    {
        \label{chart:compresses}
        \includegraphics[width=.23\textwidth]{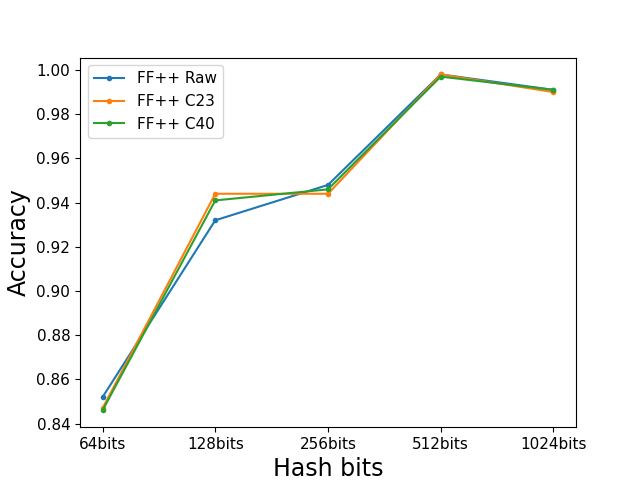}
    }
    \subfloat[Activation functions]
    {
        \label{chart:acts}
        \includegraphics[width=.23\textwidth]{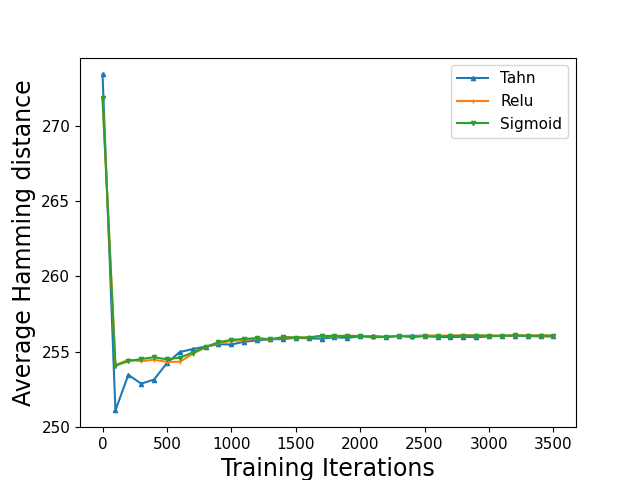}
    }
    \subfloat[Different losses]
    {
        \label{chart:hashloss}
        \includegraphics[width=.23\textwidth]{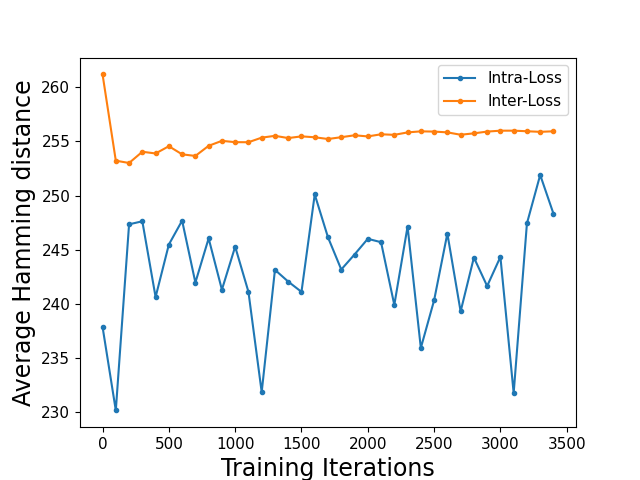}
    }
    \caption{
        (a) Tracing accuracy of different image processing. Various common video processing methods have no significant impact on accuracy except video cropping. It may be that video cropping has a great impact on video content.
        (b) Tracing accuracy of different hash bits. The accuracy increases with the hash bits. When the hash bit is 512, the accuracy is the highest.
        (c) Average Hamming distance of different activation functions. We evaluate the influence of different activation functions that the Hash Triplet Loss can keep the average Hamming distance near $\frac{k}{2}$ for all kinds of activation functions, where k is the hash bits.
        (d) Average Hamming distance with only one loss. The Inter-class loss ensures that the average Hamming distance is around $\frac{k}{2}$, and the Intra-class loss makes the hash codes of the fake video and the original video as similar as possible.
    }
    \label{fig:line_chart}
    \end{center}
\end{figure*}

\paragraph{Robustness Experiments}
There are various types of video interference processing on the Internet.
We tested several methods of video interference: image detail enhancement, Gaussian blur, image blur, median filter and cropping.
It is used to verify the anti-interference performance of ViTHash.
As shown in Table~\ref{tbl:hashbits} and Fig.~\ref{fig:line_chart}, the results of the experiments shown that there is no obvious difference between the augmented videos and the original videos.
Experiments have shown that the accuracy is higher with the increase of hash bits.
When the hash bits exceed 1024, the accuracy improvement is limited.

\paragraph{Evaluations of Cross-Dataset}
As shown in Table~\ref{tbl:cross-db-ff}, we compared the performance of cross-dataset with several recent works on FaceForensics++ (FF++)~\cite{roessler2019faceforensicspp} on five DeepFake datasets: DeepFakes (DF), Face2Face (F2F), FaceSwap (FS), NeuralTextures (NT) and FaceShifter (FSh). The experiments of cross-dataset on FaceForensics++ is trained on one dataset and tested on other datasets. Compared with recent works, we have better or equal performance on same-datasets, and have huge advantages on cross-dataset. Experiments have shown that ViTHash still works well with unknown fake videos.

\paragraph{Evaluations of Within-Dataset}
As shown in Table~\ref{tab:other-dbs}, we have done experiments on several public datasets.
On the DeepFakeDetection dataset, we training on the C23 dataset and test on the C40 dataset.
On the Celeb-DF dataset, two fake videos of each original video are used as the test set, and the rest of the fake videos are used as the training set.
On the FaceForensics++ dataset, we training on DeepFakes, Face2Face, FaceSwap and Neural textures, and test on FaceShifter.

\subsection{Evaluations of Object-Based Forgery}
\label{subsec:measurement}

As shown in Table~\ref{tab:further}, we evaluate ViTHash on three different fake datasets: DFTL, VSTL and DAVIS2016-TL, and achieved almost $100\%$ accuracy. The mIoU of DMAC and DMVN are directly cited from DMAC \cite{DMAC}.
We also evaluate Localizator on DeepFakeDetection, DFTL, VSTL and DAVIS2016-TL, and achieves the comparable performance on different fake scenes.
Experiments shown that ViTHash is effective on different fake videos, and performance of the Localizator has obvious advantages over DMAC~\cite{DMAC} and DMVN~\cite{DMVN} that may benefit from the temporal features of videos.
As shown in Fig.~\ref{fig:l_a}, Fig.~\ref{fig:l_b} and Fig.~\ref{fig:l_c}, our method is effectively for various forgery methods: DeepFake, video splicing and video object inpainting.

\begin{table}[t]
    \centering
    \caption{Evaluation on different datasets}
    \label{tab:other-dbs}
    \begin{tabular}{|c|c|c|c|}
        \cline{2-4}
        \multicolumn{1}{l|}{}                                           & DFD C40 & Celeb-DF & FF++ Raw  \\
        \hline
        ACC                                                             & 0.963   & 0.994    & 0.998     \\
        \hline
        \begin{tabular}[c]{@{}c@{}}Total\\ Original Videos\end{tabular} & 393     & 590      & 1000      \\
        \hline
        \begin{tabular}[c]{@{}c@{}}Total\\ Fake Videos\end{tabular}     & 3,068   & 5,639    & 5,000     \\
        \hline
    \end{tabular}
\end{table}

\begin{table}[t]
\centering
\caption{Results of reliability experiments}
\label{tab:further}
\begin{tabular}{c|ccl|lc} 
\toprule
\multirow{2}{*}{\textbf{Dataset}} & \multicolumn{3}{c|}{\begin{tabular}[c]{@{}c@{}}\textbf{Localization~}\\\textbf{(mIoU)}\end{tabular}} & \multicolumn{2}{c}{\begin{tabular}[c]{@{}c@{}}\textbf{\textbf{Classification}}\\\textbf{\textbf{(\%)}}\end{tabular}}  \\ 
\cline{2-6}
                                  & \multicolumn{1}{c|}{DMAC} & \multicolumn{1}{l|}{DMVN} & \multicolumn{1}{c|}{Ours}                   & \multicolumn{1}{c|}{ACC} & \begin{tabular}[c]{@{}c@{}}Total \\Originals\end{tabular}                               \\
\hline
DFD                               & -                         & -                         & 0.726                                       & 94.9                     & 363                                                                                        \\
DFTL                              & -                         & -                         & 0.880                                       & 99.9                     & 133                                                                                        \\
VSTL                              & 0.828                     & 0.751                     & \textbf{0.842}                              & 99.9                     & 30                                                                                         \\
DAVIS2016-TL                             & 0.828                     & 0.751                     & \textbf{0.882}                              & 99.9                     & 50                                                                                         \\
\bottomrule
\end{tabular}
\end{table}

\begin{table}
\centering
\setlength{\extrarowheight}{0pt}
\addtolength{\extrarowheight}{\aboverulesep}
\addtolength{\extrarowheight}{\belowrulesep}
\setlength{\aboverulesep}{0pt}
\setlength{\belowrulesep}{0pt}
\caption{Cross-dataset evaluation on FaceForensics++ and compared with recent works}
\label{tbl:cross-db-ff}
\begin{tabular}{c|l|lllll} 
\toprule
\multirow{2}{*}{\begin{tabular}[c]{@{}c@{}}Training\\Set\end{tabular}} & \multicolumn{1}{c|}{\multirow{2}{*}{Model}} & \multicolumn{5}{c}{Test Set (ACC)}                                                                                                                                                                                                             \\ 
\cline{3-7}
\multicolumn{1}{l|}{}                              & \multicolumn{1}{c|}{}                       & \multicolumn{1}{c|}{DF}                            & \multicolumn{1}{c|}{F2F}                           & \multicolumn{1}{c|}{FS}                            & \multicolumn{1}{c|}{NT}                            & \multicolumn{1}{c}{FSh}  \\ 
\hline
\multirow{4}{*}{DF}                                & Xception                                    & {\cellcolor[rgb]{0.937,0.937,0.937}}\textbf{0.993} & 0.736                                              & 0.490                                              & 0.736                                              & -                        \\
                                                   & Face X-ray\cite{Face_X-Ray}                                  & {\cellcolor[rgb]{0.937,0.937,0.937}}0.987          & 0.633                                              & 0.600                                              & 0.698                                              & -                        \\
                                                   & Grad-CAM\cite{Grad-CAM}                                    & {\cellcolor[rgb]{0.937,0.937,0.937}}0.992          & 0.764                                              & 0.497                                              & 0.814                                              & -                        \\ 
\hhline{~------}
                                                   & \textbf{ViTHash}                            & {\cellcolor[rgb]{0.937,0.937,0.937}}0.988          & \textbf{0.988}                                     & \textbf{0.988}                                     & \textbf{0.991}                                     & 0.986                    \\ 
\hline
\multirow{4}{*}{F2F}                               & Xception                                    & 0.803                                              & {\cellcolor[rgb]{0.937,0.937,0.937}}\textbf{0.994} & 0.762                                              & 0.696                                              & -                        \\
                                                   & Face X-ray\cite{Face_X-Ray}                                  & 0.630                                              & {\cellcolor[rgb]{0.937,0.937,0.937}}0.984          & 0.938                                              & 0.945                                              & -                        \\
                                                   & Grad-CAM\cite{Grad-CAM}                                    & 0.837                                              & {\cellcolor[rgb]{0.937,0.937,0.937}}\textbf{0.994} & 0.987                                              & 0.980                                              & -                        \\ 
\hhline{~------}
                                                   & \textbf{ViTHash}                            & \textbf{0.992}                                     & {\cellcolor[rgb]{0.937,0.937,0.937}}\textbf{0.994} & \textbf{0.992}                                     & \textbf{0.992}                                     & 0.992                    \\ 
\hline
\multirow{4}{*}{FS}                                & Xception                                    & 0.664                                              & 0.888                                              & {\cellcolor[rgb]{0.937,0.937,0.937}}0.994          & 0.713                                              & -                        \\
                                                   & Face X-ray\cite{Face_X-Ray}                                  & 0.458                                              & 0.961                                              & {\cellcolor[rgb]{0.937,0.937,0.937}}0.981          & 0.957                                              & -                        \\
                                                   & Grad-CAM\cite{Grad-CAM}                                    & 0.685                                              & 0.993                                              & {\cellcolor[rgb]{0.937,0.937,0.937}}0.995          & 0.980                                              & -                        \\ 
\hhline{~------}
                                                   & \textbf{ViTHash}                            & \textbf{\textbf{0.999}}                            & \textbf{0.998}                                     & {\cellcolor[rgb]{0.937,0.937,0.937}}\textbf{0.999} & \textbf{0.998}                                     & 0.999                    \\ 
\hline
\multirow{4}{*}{NT}                                & Xception                                    & 0.799                                              & 0.813                                              & 0.731                                              & {\cellcolor[rgb]{0.937,0.937,0.937}}0.991          & -                        \\
                                                   & Face X-ray\cite{Face_X-Ray}                                  & 0.705                                              & 0.917                                              & 0.910                                              & {\cellcolor[rgb]{0.937,0.937,0.937}}0.925          & -                        \\
                                                   & Grad-CAM\cite{Grad-CAM}                                    & 0.894                                              & \textbf{0.995}                                     & \textbf{0.993}                                     & {\cellcolor[rgb]{0.937,0.937,0.937}}\textbf{0.994} & -                        \\ 
\hhline{~------}
                                                   & \textbf{ViTHash}                            & \textbf{0.993}                                     & 0.992                                              & \textbf{0.993}                                     & {\cellcolor[rgb]{0.937,0.937,0.937}}0.993          & 0.993                    \\ 
\hline
FSh                                                & \textbf{ViTHash}                            & 0.988                                              & 0.988                                              & 0.990                                              & \textbf{0.993}                                     & 0.991                    \\
\bottomrule
\end{tabular}
\end{table}
\subsection{Experiments Analysis}
\label{subsec:analysis}

\paragraph{Evaluation on Similar Videos}
We shoot some similar original videos in the DFTL dataset to verify the tracing performance.
As shown in Fig.~\ref{fig:Similar-a}, tracing results of a group male videos and such videos with similar backgrounds.
These videos are shoot in similar office scenes: A is two different fake videos of the same original video, and B and C are fake videos with different lights in similar scenes.
As shown in Fig.~\ref{fig:Similar-b}, tracing results of female videos taken from different angles in the same room.
These videos are shoot in the same bedroom: A is two different fake videos of the same original video, and B and C are fake videos from different perspectives in the bedroom.

\paragraph{Errors Analysis}
As shown in Fig.~\ref{fig:error}, the tracing error examples.
The errors in this case are also in line with people’s common sense.
We analyzed the error results and found that the following situations easily lead to tracing failure:
\begin{itemize}
    \item The videos are similar in shape or structure.
    \item The videos with similar colors over a large area.
    \item The background and pose are almost the same.
\end{itemize}

\subsection{Ablation Studies}
\label{subsec:study}

\paragraph{Intra-Class Loss}
As shown in Fig.~\ref{chart:hashloss}, when training with only Intra-Class loss.
Although we have tried a variety of algorithm improvements and different training strategies, the hashes are always volatile and approaches to $\Vec{0}$ or $\Vec{1}$.
As shown in Fig.~\ref{chart:acts}, when the two losses are trained together, the average Hamming distance of the Hash Centers gradually stabilizes and approaches half of the hash bits.

\paragraph{Inter-Class Loss}
As shown in Fig.~\ref{chart:hashloss}, when training with only the Inter-class loss.
We found that accuracy cannot be effectively improved and it is difficult to meet our expectations.

\paragraph{Various Activation Functions}
\begin{align}
    \mathcal{F}_{tahn} &=(\sign{x} + 1) / 2 \in \left \{ 0,1 \right \} ^{k}  \label{eq:tahn}\\
    \mathcal{F}_{sigmoid} &=(\sign{x - 0.5} + 1) / 2\in \left \{ 0,1 \right \} ^{k}  \label{eq:sigmoid} \\
    \mathcal{F}_{relu} &=\sign{x} \in \left \{ 0,1 \right \} ^{k} \label{eq:relu}
\end{align}
We have tried a variety of output activation functions and the corresponding hash binary functions, such as ReLU with Eq. \eqref{eq:tahn}, Tahn with Eq. \eqref{eq:sigmoid} and Sigmoid with Eq. \eqref{eq:relu}. As shown in Fig.~\ref{chart:acts}, with the help of Hash Triplet Loss, the Hamming distance of these can be fast and stable at half of the hash bits.

\section{Conclusions}
In this paper, we use video hashing retrieval to trace the source of fake videos. Our proposed method is named ViTHash, which can provide convincing evidence. In addition, we propose a novel hash loss named Hash Triplet Loss, which makes the quality of trained hash centers like pre-trained hashes (Near-Optimal hashes). The Hash Triplet Loss explored the similarities of the intrinsic data structure by adaptively constructing the hash centers. The Hash Triplet Loss slowly lessens the Hamming distance of intra-classes and enlarges the Hamming distance of inter-classes. As a result, the average Hamming distance of hash centers is half of the hash bits. Compared with existing fake video detection methods, the ViTHash does not depend on the method of forgery and shape of objects in the video, so it can be easily extended to other forgery detection.

\section*{Data Availability}
The data used to support the findings of this study are included within the article.
\section*{Conflict of Interest}
The authors declare that they have no conflicts of interest to this work.
 
\section*{Acknowledgments}
This work was supported by Natural Science Foundation of China (NSFC) under 61972390, 61902391
and 61872356, and National Key Technology Research and Development
Program under 2019QY0701, 2019QY2202 and 2020AAA0140000. We would like to thank Zewen Long and Xiaowei Yi for the help with the preparation of experimental materials.

    {
        \bibliographystyle{IEEEtran}
        \bibliography{IEEEabrv,main}
    }

\end{document}